\documentclass[lettersize,journal]{IEEEtran}
\usepackage{stfloats}
\usepackage{cite}
\usepackage{amsmath,amssymb,amsfonts,amsthm}
\usepackage[pdftex]{graphicx}
\usepackage{textcomp}
\usepackage[pdftex]{xcolor}

\usepackage{latexsym}
\usepackage{comment}
\usepackage{algpseudocode}
\usepackage{algorithmicx, algorithm}
\usepackage{subfigure}
\usepackage{url}
\usepackage{bm}
\usepackage{booktabs}
\usepackage{here}
\usepackage{multirow}
\usepackage{mathtools}
\usepackage{autobreak}
\usepackage{adjustbox}
\usepackage{array}

\allowdisplaybreaks[3]

\makeatletter
\newcommand*\rel@kern[1]{\kern#1\dimexpr\macc@kerna}
\newcommand*\widebar[1]{%
  \begingroup
  \def\mathaccent##1##2{%
    \rel@kern{0.8}%
    \overline{\rel@kern{-0.8}\macc@nucleus\rel@kern{0.2}}%
    \rel@kern{-0.2}%
  }%
  \macc@depth\@ne
  \let\math@bgroup\@empty \let\math@egroup\macc@set@skewchar
  \mathsurround\z@ \frozen@everymath{\mathgroup\macc@group\relax}%
  \macc@set@skewchar\relax
  \let\mathaccentV\macc@nested@a
  \macc@nested@a\relax111{#1}%
  \endgroup
}
\makeatother

\usepackage{scalerel,stackengine}
\stackMath
\newcommand\widecheck[1]{%
\savestack{\tmpbox}{\stretchto{%
  \scaleto{%
    \scalerel*[\widthof{\ensuremath{#1}}]{\kern-.6pt\bigwedge\kern-.6pt}%
    {\rule[-\textheight/2]{1ex}{\textheight}}
  }{\textheight}%
}{0.5ex}}%
\stackon[1pt]{#1}{\scalebox{-1}{\tmpbox}}%
}
\def\BibTeX{{\rm B\kern-.05em{\sc i\kern-.025em b}\kern-.08em
    T\kern-.1667em\lower.7ex\hbox{E}\kern-.125emX}}

\newtheorem{thm}{Theorem}
\newtheorem{cor}{Corollary}

\newtheorem{asm}{Assumption}

\newcommand{\argmax}{\mathop{\rm arg~max}\limits}
\newcommand{\argmin}{\mathop{\rm arg~min}\limits}
\newcommand{\indep}{\mathop{\perp\!\!\!\!\perp}} 

\algnewcommand{\Inputs}[1]{
      \State \textbf{Inputs:}
      \State \parbox[t]{\linewidth}{\raggedright #1}
}
\algnewcommand{\Outputs}[1]{
      \State \textbf{Outputs:}
      \State \parbox[t]{\linewidth}{\raggedright #1}
}
\algnewcommand{\Process}[1]{
      \State \textbf{Process:}
      \State \parbox[t]{\linewidth}{\raggedright #1}
}

\begin{document}

\title{Learning from Complementary Features}

\author{Kosuke Sugiyama and Masato Uchida, {\it Member, IEEE}}

\markboth{}%
{Shell \MakeLowercase{\textit{et al.}}: A Sample Article Using IEEEtran.cls for IEEE Journals}

\IEEEpubid{}

\maketitle

\begin{abstract}
While precise data observation is essential for the learning processes of predictive models, it can be challenging owing to factors such as insufficient observation accuracy, high collection costs, and privacy constraints.
In this paper, we examines cases where some qualitative features are unavailable as precise information indicating ``what it is,'' but rather as complementary information indicating ``what it is not.''
We refer to features defined by precise information as ordinary features (OFs) and those defined by complementary information as complementary features (CFs).
We then formulate a new learning scenario termed Complementary Feature Learning (CFL), where predictive models are constructed using instances consisting of OFs and CFs.
The simplest formalization of CFL applies conventional supervised learning directly using the observed values of CFs.
However, this approach does not resolve the ambiguity associated with CFs, making learning challenging and complicating the interpretation of the predictive model's specific predictions.
Therefore, we derive an objective function from an information-theoretic perspective to estimate the OF values corresponding to CFs and to predict output labels based on these estimations.
Based on this objective function, we propose a theoretically guaranteed graph-based estimation method along with its practical approximation, for estimating OF values corresponding to CFs.
The results of numerical experiments conducted with real-world data demonstrate that our proposed method effectively estimates OF values corresponding to CFs and predicts output labels.

\end{abstract}


\section{Introduction}
\label{sec:intro}

Precise data observation is essential for constructing predictive models using supervised learning.
However, obtaining precise data can be challenging due to insufficient observation accuracy, high collection costs, and privacy constraints.
Consequently, various methods have been proposed to facilitate learning in situations where precise data are unavailable.
Regarding output labels, several approaches address different scenarios: semi-supervised learning \cite{chapelle2006semi} handles cases where some instances have missing output labels, noisy label learning \cite{natarajan2013learning} deals with the possibility of incorrect label assignments, partial label learning \cite{cour2011learning} addresses situations where only a subset of labels containing the true label is provided,
and complementary label learning \cite{ishida2017cll,ishida2019complementary} is designed for cases where only information indicating that a label is not the true label is available.
For input features, the impute-then-regress approach \cite{morvan2020neumiss,morvan2021whats} has been proposed to address situations where some input features may be missing for each instance.

In this study, we focus on predictive performance when only complementary input information indicating ``what it is not'' is available, rather than precise input information indicating ``what it is.''
We refer to qualitative features defined by complementary information as {\it complementary features (CFs)}, and designate this learning scenario as {\it Complementary Feature Learning (CFL)}.
Even when precise values for certain features are difficult to obtain due to constraints such as observation accuracy, collection costs, or privacy, complementary information can often be more accessible.
For instance, when multiple candidates exist for a feature's value, selecting a specific value among them can enhance observation accuracy but also incurs additional costs. 
However, since the values not included among the candidates are already known, it is straightforward to identify what the feature value is not. 
This complementary information can be utilized as a CF's value, thereby avoiding additional costs.

\begin{figure*}[t]
    \centering
    \includegraphics[width=0.8\linewidth]{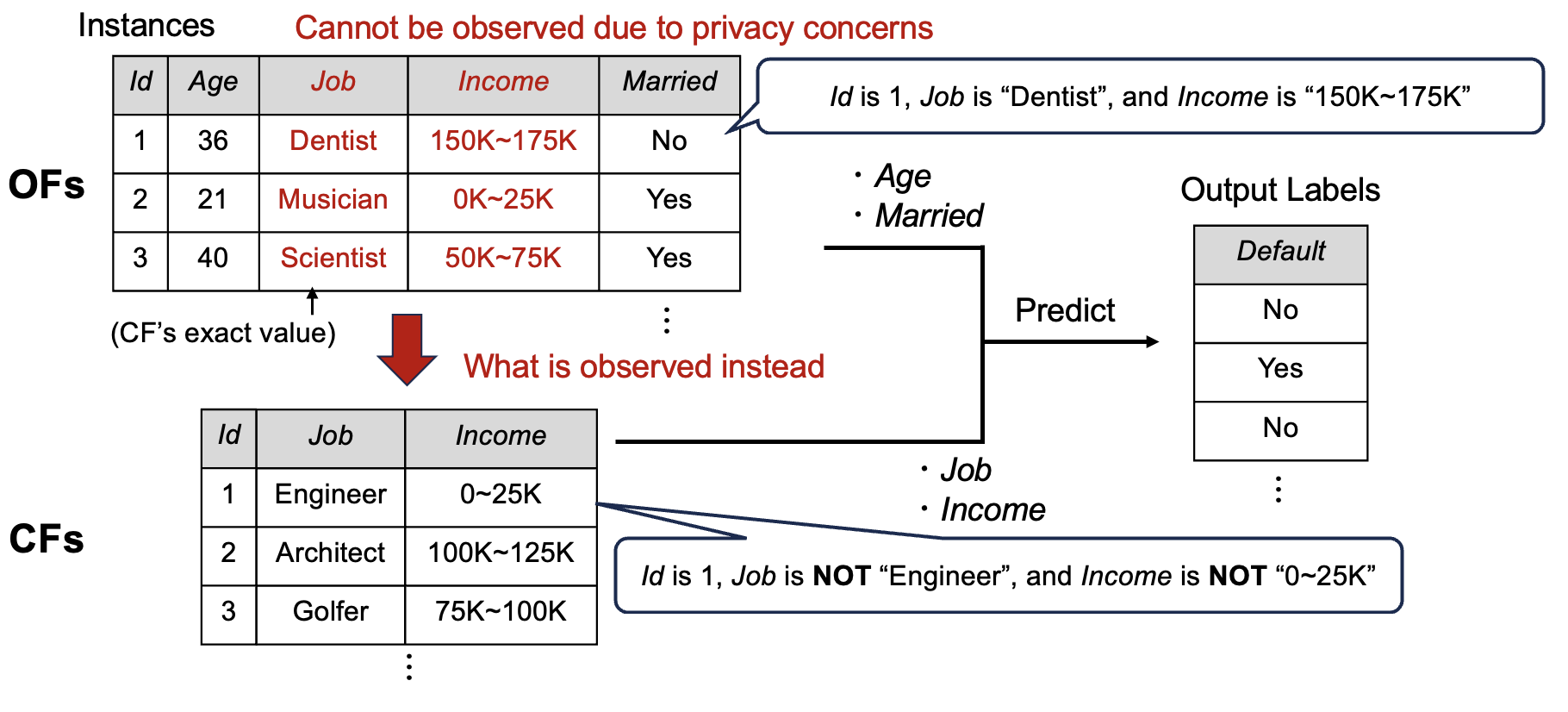}
    \caption{Example of CFL problem setting.
             The problem involves predicting a binary output label {\it Default} based on four features: {\it Age}, {\it Married}, {\it Job}, and {\it Income}. 
             Due to privacy concerns, the OF's values of {\it Job} and {\it Income} cannot be observed; instad, the CF's values of these features are observed.
             }
    \label{fig:CF_overview}
\end{figure*}

In this problem setting, there are two major challenges.
Firstly, the direct use of CFs in learning presents challenges in training predictive models, as the value of the original feature corresponding to a CF (hereinafter referred to as the {\it exact value} of the CF) is ambiguous.
Secondly, the interpretability of predictive labels generated by the predictive model may be reduced, since explanations of the predictive model's outputs based on CFs are inherently more difficult to interpret than those based on ordinary features (OFs) that are not CFs.
Figure \ref{fig:CF_overview} illustrates this specific situation.
The most straightforward approach to address this challenge is to estimate the exact values of CFs and incorporate these estimates into the learning process.

As the first contribution of this study, we derive an objective function from an information-theoretic perspective to predict output labels using ordinary supervised learning based on the estimation of CFs' exact values.
This objective function is naturally derived from the one used in ordinary supervised learning.
Specifically, we demonstrate that the upper bound of the predictive loss measured by Kullback-Leibler (KL) divergence in ordinary supervised learning is represented by the sum of terms evaluating the estimation quality of CFs' exact values using mutual information and the predictive loss measured by KL divergence when using the estimation results of CFs' exact values.
We adopt this upper bound as the objective function for CFL and justify a corresponding learning method that consists of two steps: estimating the CFs' exact values and using these estimates in ordinary supervised learning to predict output labels.

As the second contribution of this study, we propose an iterative graph-based method for estimating the CFs' exact values under the aforementioned objective function.
First, we derive a method with theoretically guaranteed effectiveness under the hypothetical scenario where the CFs' exact values can be self-referenced, meaning the estimation is conducted by referencing the values of the entities being estimated.
This method aims to estimate the confidence representing the predictive probability of the exact values of CFs for each instance and iteratively updates the confidences by calculating the weighted average of the confidences of all instances.
We prove that, at each iteration, the estimated confidences approach the probability distribution of the CFs' exact values.
However, two challenges arise with the confidence update process.
The first challenge is that weight optimization cannot be executed in practice because the exact values of CFs must be known for this step.
The second challenge is that the computational complexity of the confidence update increases exponentially with the number of CFs, making execution difficult as the number of CFs increases.

To address the first challenge, we apply the smoothness assumption \cite{chapelle2006semi}, which posits that points similar in the input space are also similar in the output space.
This allows us to substitute the optimization problem involving the exact values of CFs with one that uses OFs.
For the second challenge, we substitute confidence updates for all CFs with more computationally efficient confidence updates for each CF and demonstrate the theoretical validity of this approach. 
Under these substitutes, we formulate an approximately computable method for the aforementioned theoretically guaranteed method.
This method can be interpreted as estimating the exact values of CFs by constructing a similarity graph, where nodes represent instances and edge weights indicate their similarities, and propagating the confidence in the CFs' exact values over this graph.
Our method is inspired by weakly supervised learning methods \cite{zhu2002learning,zhang2015solving}, which address scenarios in which output labels are missing for some samples or where sets containing true labels are provided as output labels.
The effectiveness of our proposed method was confirmed through numerical experiments using real-world data.
Our main contributions can be summarized as follows:

\begin{enumerate}
    \item We demonstrated that the objective function for performing a learning task estimating CFs' exact values and predicting output labels based on these estimations using ordinary supervised learning can be naturally derived information-theoretically from the formulation of ordinary supervised learning.
    Based on the derivation, we formulated CFL.
    \item We proposed a graph-based iterative method for estimating CFs' exact values under the derived objective function.
    We derived this method as a practical approximation of a theoretically validated method by applying the smoothness assumption.
    Additionally, we introduced a strategy to prevent the computational complexity of this approximation method from increasing exponentially with respect to the number of CFs.
    \item Through numerical experiments using real-world data, we confirmed that our proposed method not only accurately estimates CFs' exact values but also significantly enhances the prediction performance of output labels.
\end{enumerate}

\section{Related Works}
\label{sec:related_works}

When estimating the exact values of CFs, one can apply a form of weakly supervised learning known as complementary label learning (CLL) \cite{ishida2017cll}.
CLL is a methodology designed for situations where complementary labels (CLs), which indicate ``this is not a true label,'' are provided as output labels \cite{ishida2017cll,ishida2019complementary}.
By treating CF values as output labels, the learning problem aligns with the CLL framework for predicting CF values.
Ishida et al. derived an unbiased estimator of risk in ordinary supervised learning that can be applied to data with assigned CLs.
They proposed a CLL method that utilizes this estimator as the objective function \cite{ishida2017cll,ishida2019complementary}.
These learning methods primarily employ deep learning models.
Another learning strategy has been proposed to facilitate the application of learning algorithms such as gradient boosting with log loss \cite{lin2023reduction}.
Therefore, various learning algorithms commonly used in ordinary supervised learning can be applied when estimating the exact values of the CFs based on CLL.

Similarly to CLL, partial label learning (PLL) \cite{cour2011learning} can also be employed for estimating the exact values of CFs.
PLL is a learning methodology that uses partial labels (PLs) \cite{cour2011learning}, which are defined as a set containing the true label treated as the output label.
When the total number of labels is $K$ and the number of elements in a PL is $K-1$, the PL implies that the label not included in this set is not the true label, effectively functioning as a CL.
Because PLs can generalize CLs \cite{katsura20bridging}, PLL is applicable in the context of this study.

PLL methods are categorized into two types based on whether they are used to construct predictive models.
Methods used to construct predictive models include those that treat all labels assigned as PLs as true labels when solving prediction problems \cite{cour2011learning}, and those that regard true labels as latent variables to be estimated \cite{jin2002learning,nguyen-caruana2008classfication}.
In contrast, methods not used to construct predictive models typically employ graph-based estimation techniques \cite{gong-liu2018regularization,sun2020pp,zhang2015solving}.
These methods create a similarity graph between input instances and estimate true labels based on this graph.
There are primarily two methods for estimating true labels using this graph.
The first method solves optimization problems to determine confidences associated with true labels \cite{gong-liu2018regularization,sun2020pp}.
The second method propagates confidences associated with true labels based on PLs over the graph to estimate the true labels \cite{zhang2015solving}.
Methods based on confidence propagation, like this second method, have also been proposed for semi-supervised learning (SSL), which learns in scenarios where unsupervised instances are included in the training data \cite{zhu2002learning,zhou2003label-spreading}.
Label propagation \cite{zhu2002learning} is a representative method of confidence propagation.
This method propagates confidences associated with true labels from supervised instances to unsupervised instances.

In principle, all the aforementioned methods can be used to estimate the exact values of CFs.
However, applying methods proposed in CLL or PLL for constructing predictive models may be impractical for this task due to significant computational costs.
This is because it is necessary to solve multiple prediction problems for various CFs before addressing the primary objective of predicting output labels.
Furthermore, each prediction problem requires not only the construction of predictive models but also feature engineering tailored to the prediction of each CF.

Accordingly, we focus on methods that estimate true labels without constructing predictive models.
One such method is {\it Instance-based PArtial Label learning} (IPAL) \cite{zhang2015solving}, a graph-based estimation method within PLL using confidence propagation.
The main process of this method can be summarized in two steps.
In the first step, a similarity graph is constructed with instances as nodes in the input space.
The weights of the edges are determined by approximating each instance through a linear combination of its adjacent instances.
In the second step, true labels are estimated by iteratively propagating their confidences based on PLs over the similarity graph created in the first step.
Label propagation \cite{zhu2002learning} and label spreading \cite{zhou2003label-spreading}, which employ similar confidence propagation methods, have also been validated in SSL.
The main distinction between these methods and IPAL lies in the computation of edge weights
IPAL determines them based on the aforementioned approximation of each instance, whereas label propagation and label spreading determine them using a Gaussian kernel.

When applying a method based on confidence propagation to estimate the exact values of CFs, constructing a similarity graph, only needs to be performed once for all CFs.
Furthermore, since confidence propagation is executed in exactly the same manner for all CFs, there is no need to optimize the method for each CF individually.
Consequently, it is relatively straightforward to apply confidence propagation methods to estimate the exact values of CFs.
However, when applying these methods to CFL, two issues arise: first, the application of these methods under the CFL formulation described later is not always theoretically justified; second, they do not utilize the relationships between multiple CFs.
Therefore, in this study, we propose a new confidence propagation method for CFL that addresses these issues.

Our proposed method has two primary characteristics.
Firstly, it is derived as an approximation of a theoretically guaranteed method based on the CFL formulation.
Secondly, our method leverages the presence of multiple CFs, for which exact values need to be estimated.
To utilize the relationship between CFs, we consider the estimation results of CFs' exact values as pseudo-exact values and use them to apply the confidence propagation algorithm.

\section{Formulation of CFL}
\label{sec:formulation}

In this section, we formalize CFL.
Section \ref{subsec:notation} describes all relevant notations.
In Section \ref{subsec:derivation_obj_func}, the objective function of CFL is derived from an information-theoretic perspective.
This derived objective function supports the strategy of estimating the exact values of CFs and using these estimates to predict output labels.

\subsection{Notation}
\label{subsec:notation}

In this paper, the notation $[n],~n \in \mathbb{N}^+$ represents the set $\{1,\dots,n\}$. 
Let $F^{c}$ denote the number of CFs, and $F^{o}$ denote the number of OFs.
Let $X_{j}^{c}$ be the random variable representing the exact value of the $j$-th feature, ($j \in [F^{c}]$) observed as a CF, and $\mathcal{X}_{j}^{c}$ be the set of its possible values. 
Let $\widebar{X}_{j}^{c}$ be the random variable representing the observed value of the $j$-th feature, ($j \in [F^{c}]$) observed as a CF, and $\widebar{\mathcal{X}}_{j}^{c}$ be the set of its possible values.
Throughout this study, each CF is assumed to have a single value that is distinct from its exact value.
Therefore, $\widebar{\mathcal{X}}^{c}_{j} \equiv \mathcal{X}^{c}_{j}$.
For simplicity, but without loss of generality, we specifically assume $\mathcal{X}_{j}^{c} = \{1,\dots,u^{c}_j\}$, where $u^{c}_j$ represents the number of possible values that $X_{j}^{c}$ can take. 
Let $X_{j}^{o}$ be the random variable representing the observed value of the $j$-th feature, ($j \in [F^{o}]$) observed as an OF, and $\mathcal{X}_{j}^{o}$ be the set of its possible values.
The input spaces of CFs, OFs, and the entire feature set are denoted by $\mathcal{X}^{c} = \prod_{j \in [F^{c}]}\mathcal{X}^{c}_{j}$, $\mathcal{X}^{o} = \prod_{j \in [F^{o}]}\mathcal{X}^{o}_{j}$, and $\mathcal{X}=\mathcal{X}^{c} \times \mathcal{X}^{o}$, respectively.
We denote the random variable vector of CFs by $\widebar{\bm{X}}^{c} = (\widebar{X}_{1}^{c}, \ldots, \widebar{X}_{F^{c}}^{c}) \in \widebar{\mathcal{X}}^{c}~(\equiv\mathcal{X}^{c})$ and that of OFs by $\bm{X}^{o} = (X_{1}^{o}, \ldots, X_{F^{o}}^{o}) \in \mathcal{X}^{o}$.
Furthermore, we denote the vector of random variables for exact values of $\widebar{\bm{X}}^{c}$ as $\bm{X}^{c} = (X_{1}^{c}, \ldots, X_{F^{c}}^{c}) \in \mathcal{X}^{c}~ (\equiv\widebar{\mathcal{X}}^{c})$.
Let $\bm{X}=(\bm{X}^{c}, \bm{X}^{o})$ be a random variable vector consisting only of the exact values of input features, and $\widebar{\bm{X}}=(\widebar{\bm{X}}^{c}, \bm{X}^{o})$ be a random variable vector of input features that includes CFs.
Let $Y$ denote the random variable representing the output label.

\begin{figure}[t]
    \centering
    \includegraphics[width=0.6\linewidth]{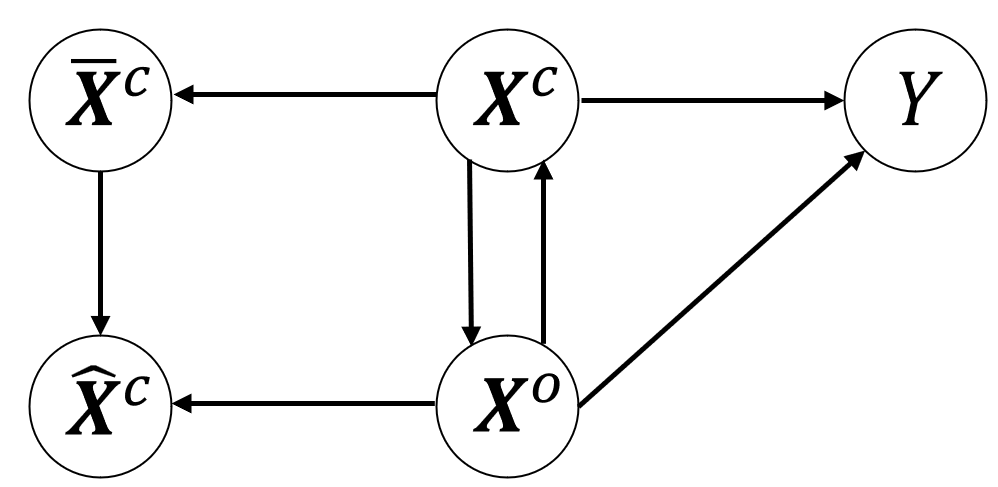}
    \caption{Graphical model showing dependencies between variables}
    \label{fig:graphical_model}
\end{figure}

Let $p_*(\bm{x},y)$ denote the true distribution of $\bm{X}$ and $Y$.
We assume that $\widebar{\bm{X}}^{c}$ depends only on $\bm{X}^{c}$, which can be represented as $\bar{p}(\bar{\bm{x}}^{c} | \bm{x}^c)$.
Let $\{(\bar{\bm{x}}^{c}_i, \bm{x}^{o}_i, y_i)\}_{i\in [n]}$ denote the training data obtained and $\{\bm{x}^{c}_i\}_{i\in [n]}$ denote the exact values corresponding to CFs for the training instances, where $n$ is the size of the training set.
We assume that each sample $(\bar{\bm{x}}^{c}_i, \bm{x}^{o}_i, y_i, \bm{x}^{c}_i)$ is i.i.d. drawn from $\bar{p}(\bar{\bm{x}}^{c} | \bm{x}^c)p_*(\bm{x},y)$.
For any $i \in [n]$, the vectors $\bar{\bm{x}}^{c}_i$, $\bm{x}^{o}_i$, and $\bm{x}^{c}_i$ are specifically represented as $\bar{\bm{x}}^{c}_{i}=(\bar{x}^{c}_{i1},\dots,\bar{x}^{c}_{i F^{c}})$, $\bm{x}^{o}_{i}=(x^{o}_{i1},\dots,x^{o}_{i F^{o}})$, and $\bm{x}^{c}_{i}=(x^{c}_{i1},\dots,x^{c}_{i F^{c}})$, respectively.
All predictive models used throughout this study are represented by probability density functions.
The function that predicts $Y$ from $\bm{X}$ is referred to as the {\it label prediction model} and denoted by $p_{\bm{\theta}}(y|\bm{x}^{c}, \bm{x}^{o})$, where $\bm{\theta}$ represents its parameters.
The function that estimates $\bm{X}^c$ from $\widebar{\bm{X}}^{c}$ and $\bm{X}^{o}$ is called the {\it feature estimation model} and denoted by $q_{\bm{\eta}}(\bm{x}^{c} | \bar{\bm{x}}^{c}, \bm{x}^{o})$, with $\bm{\eta}$ representing its parameters.
The random variable following $q_{\bm{\eta}}(\bm{x}^{c} | \bar{\bm{x}}^{c}, \bm{x}^{o})$ is denoted by $\widehat{\bm{X}}^c$.
Figure \ref{fig:graphical_model} illustrates the dependencies among these random variables.
The two directed edges connecting $\bm{X}^c$ and $\bm{X}^{o}$ indicate possible bidirectional dependencies between these random variables in their combinations.
Although this discussion primarily focuses on classification problems, the same principles is applicable to regression problems without loss of generality.

\subsection{Derivation of the Objective Function}
\label{subsec:derivation_obj_func}

In this section, we derive the objective function for CFL from the objective function for ordinary supervised learning with exact values of all features, based on an information-theoretic perspective.
The strategy of estimating the exact values of CFs and using them to predict output labels is justified by the derived objective function.

The objective function used to construct the label prediction model $p_{\bm{\theta}}$ in ordinary supervised learning is expressed naturally using KL divergence, as follows:
\begin{align}
\begin{split}
    &D_{\mathrm{KL}}(p_*(Y|\bm{X}^{c},\bm{X}^{o}) || p_{\bm{\theta}}(Y|\bm{X}^{c}, \bm{X}^{o})) \\ 
    &~~~~~~~~~~ = \mathbb{E}_{p_*(\bm{x}^{c}, \bm{x}^{o}, y)} \bigg[ \log \frac{p_*(y|\bm{x}^{c},\bm{x}^{o})}{p_{\bm{\theta}}(y|\bm{x}^{c}, \bm{x}^{o})} \bigg].
\end{split}
\label{eq:ordinay_KL}
\end{align}
However, since $\bm{X}^c$ is not available in CFL, we cannot directly evaluate Eq. \eqref{eq:ordinay_KL}.
Instead, we substitute the estimated values $\widehat{\bm{X}}^c$, obtained from the feature estimation model $q_{\bm{\eta}}$ with the CFs' exact values $\bm{X}^c$.
We now consider the following function as the objective function for learning $p_{\bm{\theta}}$:
\begin{align}
\begin{split}
    &D_{\mathrm{KL}}(p_*(Y|\widehat{\bm{X}}^{c},\bm{X}^{o}) || p_{\bm{\theta}}(Y|\widehat{\bm{X}}^{c}, \bm{X}^{o}))  \\
      &~~~~~~~~~~ =\mathbb{E}_{q_{\bm{\eta}}(\hat{\bm{x}}^c|\bm{x}^{o})p_*(\bm{x}^{o},y)} \bigg[ \log \frac{p_*(y|\hat{\bm{x}}^{c},\bm{x}^{o})}{p_{\bm{\theta}}(y|\hat{\bm{x}}^{c}, \bm{x}^{o})} \bigg].
\end{split}  
\label{eq:CFL_KL}
\end{align}
Here, $q_{\bm{\eta}}(\hat{\bm{x}}^c|\bm{x}^{o}) =\mathbb{E}_{\bar{p}(\bar{\bm{x}}^{c}|\bm{x}^c) p_*(\bm{x}^c|\bm{x}^{o})}[q_{\bm{\eta}}(\hat{\bm{x}}^{c} | \bar{\bm{x}}^{c}, \bm{x}^{o})]$.
The following relationship between Eqs. \eqref{eq:ordinay_KL} and \eqref{eq:CFL_KL} holds.
A proof is given in Appendix \ref{subsec:proof_obj_relation}.

\begin{thm}\label{thm:objective_relation}
    \begin{align}
    \begin{split}
         &D_{\mathrm{KL}}(p_*(Y|\bm{X}^{c},\bm{X}^{o}) || p_{\bm{\theta}}(Y|\bm{X}^{c}, \bm{X}^{o})) \\
         &\le D_{\mathrm{KL}}(p_*(Y|\widehat{\bm{X}}^{c},\bm{X}^{o}) || p_{\bm{\theta}}(Y|\widehat{\bm{X}}^{c}, \bm{X}^{o})) \\
         &~~~~~~~~~~~~~~~~~~~~~ + (\mathbb{I}(Y,\bm{X}^{c}|\bm{X}^{o}) - \mathbb{I}(Y, \widehat{\bm{X}}^c | \bm{X}^{o}))
    \end{split}
    \label{eq:objective_relation}
    \end{align}
Here, $\mathbb{I}(\cdot|\cdot)$ represents the conditional mutual information. 
$\mathbb{I}(Y,\bm{X}^{c}|\bm{X}^{o})$ and $\mathbb{I}(Y, \widehat{\bm{X}}^c | \bm{X}^{o})$ are defined as follows:
\begin{align}
  &\mathbb{I}(Y,\bm{X}^{c}|\bm{X}^{o}) = \mathbb{E}_{p_*(\bm{x},y)} \bigg[\log \frac{p_*(y,\bm{x}^{c} |\bm{x}^{o})}{p_*(y|\bm{x}^{o})p_*(\bm{x}^{c} |\bm{x}^{o})} \bigg], \label{eq:MI_star} \\
  &\mathbb{I}(Y, \widehat{\bm{X}}^c | \bm{X}^{o}) = \mathbb{E}_{q_{\bm{\eta}}(\hat{\bm{x}}^{c}|\bm{x}^{o})p_*(\bm{x}^{o},y)}\bigg[\log \frac{p_{*,\bm{\eta}}(y, \hat{\bm{x}}^{c}|\bm{x}^{o})}{p_*(y|\bm{x}^{o})q_{\bm{\eta}}(\hat{\bm{x}}^{c}|\bm{x}^{o})}  \bigg]. \label{eq:MI_hat}
\end{align}

Here, $p_{*,\bm{\eta}}(y, \hat{\bm{x}}^{c}|\bm{x}^{o}) = \mathbb{E}_{\bar{p}(\bar{\bm{x}}^{c}|\bm{x}^{c}) p_*(\bm{x}^{c}|\bm{x}^{o})}[ p_*(y|\bm{x}^{c},\bm{x}^{o}, \bar{\bm{x}}^{c}) \\ q_{\bm{\eta}}(\hat{\bm{x}}^{c}|\bm{x}^{o}, \bar{\bm{x}}^{c}) ]$.

\end{thm}

The first term on the RHS of Eq. \eqref{eq:objective_relation}, which corresponds to Eq. \eqref{eq:CFL_KL}, evaluates the accuracy of the label prediction model $p_{\bm{\theta}}$ when using the estimated exact values of CFs.
This first term is denoted hereinafter as:
\begin{align}
    J_{\mathrm{KL}}(\bm{\theta}, \bm{\eta}) := D_{\mathrm{KL}}(p_*(Y|\widehat{\bm{X}}^{c},\bm{X}^{o}) || p_{\bm{\theta}}(Y|\widehat{\bm{X}}^{c}, \bm{X}^{o})).
\end{align}
The second term on the RHS of Eq. \eqref{eq:objective_relation} quantifies the discrepancy between the information regarding input-output relationships in ordinary learning and that obtained when the estimated result from the feature estimation model $q_{\bm{\eta}}$ is substituted for $\bm{X}^{c}$.
In other words, this term measures the extent of information loss associated with input-output relationships due to the substitution.
Therefore, this term can be interpreted as an accuracy measure of the feature estimation model $q_{\bm{\eta}}$.
This second term is denoted hereinafter as:
\begin{align}
    J_{\mathrm{MI}}(\bm{\eta}) := \mathbb{I}(Y,\bm{X}^{c}|\bm{X}^{o}) - \mathbb{I}(Y, \widehat{\bm{X}}^c | \bm{X}^{o}).
\end{align}
From the above discussion, we consider $J_{\mathrm{KL}}+J_{\mathrm{MI}}$ as the objective function for CFL, which is an upper bound on the objective function defined by Eq. \eqref{eq:ordinay_KL} for ordinary supervised learning.

It is important to note that, because $J_{\mathrm{KL}}$ depends on both $q_{\bm{\eta}}$ and $p_{\bm{\theta}}$, $\bm{\eta}$ and $\bm{\theta}$ must be simultaneously optimized when minimizing $J_{\mathrm{KL}}+J_{\mathrm{MI}}$.
One of the simplest optimization methods for $\bm{\eta}$ and $\bm{\theta}$ is through iterative optimization.
However, because $q_{\bm{\eta}}$ and $p_{\bm{\theta}}$ are typically constructed using machine learning algorithms, this iterative process incurs high computational costs, rendering it impractical.
Instead, we first optimize the feature estimation model $q_{\bm{\eta}}$ using $J_{\mathrm{MI}}$ as the objective function.
Then, we fix the optimized $q_{\bm{\eta}}$, and optimize the label prediction model $p_{\bm{\theta}}$ using $J_{\mathrm{KL}}$ as the objective function.
This approach allows $q_{\bm{\eta}}$ and $p_{\bm{\theta}}$ to be optimized individually, only once, effectively reducing computational costs and enhancing practicality.

When optimizing the feature estimation model $q_{\bm{\eta}}$ using $J_{\mathrm{MI}}$, there arises a challenge that the applicable optimization methods are limited.
When optimizing the label prediction model $p_{\bm{\theta}}$, supervised learning based on maximum likelihood estimation (or using cross-entropy as the loss function) can be employed, facilitating the use of various learning algorithms.
Representative examples include logistic regression and neural networks using cross-entropy loss. 
Conversely, during the optimization of the feature estimation model $q_{\bm{\eta}}$, the objective function $J_{\mathrm{MI}}$ prevents the application of PLL or CLL methods based on maximum likelihood estimation as it is not the negative log-likelihood.

Therefore, we aim to construct an objective function alternative to $J_{\mathrm{MI}}$ that allows for optimization of $q_{\bm{\eta}}$ using maximum likelihood estimation.
To address this requirement, we present the following theorem. 
\begin{thm}
    The following inequality holds:
    \begin{align}
    J_{\mathrm{MI}}(\bm{\eta}) = \mathbb{I}(Y,\bm{X}^{c}|\bm{X}^{o}) - \mathbb{I}(Y, \widehat{\bm{X}}^c | \bm{X}^{o}) \ge 0.
    \label{eq:J_MI_lower_bound}
    \end{align}
    In addition, the following holds:
    \begin{align}
        D_{\mathrm{KL}}(p_*(\bm{X}^{c}|\bm{X}^{o}) || q_{\bm{\eta}}(\bm{X}^{c}|\bm{X}^{o})) =0 \Rightarrow J_{\mathrm{MI}} = 0.
    \end{align}
\label{thm:J_MI}
\end{thm}
A proof is given in Appendix \ref{subsec:proof_J_MI}.
According to Theorem \ref{thm:J_MI}, when $q_{\bm{\eta}}$ is optimized to minimize $D_{\mathrm{KL}}(p_*(\bm{X}^{c}|\bm{X}^{o}) || q_{\bm{\eta}}(\bm{X}^{c}|\bm{X}^{o}))$, $J_{\mathrm{MI}}$ is also minimized. 
Additionally, minimizing $D_{\mathrm{KL}}(p_*(\bm{X}^{c}|\bm{X}^{o}) || q_{\bm{\eta}}(\bm{X}^{c}|\bm{X}^{o}))$ with respect to $q_{\bm{\eta}}$ is equivalent to minimizing the negative log-likelihood, which allows for the application of PLL or CLL methods based on maximum likelihood estimation.
Therefore, in this study, we consider optimizing the feature estimation model $q_{\bm{\eta}}$ using a surrogate objective function $J_{\mathrm{SL}}$, defined as follows:
\begin{align}
    J_{\mathrm{SL}}(\bm{\eta})\equiv D_{\mathrm{KL}}(p_*(\bm{X}^{c}|\bm{X}^{o}) || q_{\bm{\eta}}(\bm{X}^{c}|\bm{X}^{o})).
    \label{eq:obj_surrogate}
\end{align}

However, because $\bm{X}^c$ cannot be observed in practice, it is impossible to directly optimize $q_{\bm{\eta}}$ using $J_{\mathrm{SL}}$ as the objective function.
Instead, $J_{\mathrm{SL}}$ must be minimized in an approximate manner without using $\bm{X}^c$.
Therefore, as a preliminary step in developing such a minimization method, we first consider a hypothetical scenario wherein $\bm{X}^c$ can be self-referenced, meaning the estimation is conducted by referencing the values of the entities being estimated.
Under this assumption, we construct a graph-based iterative estimation method that is theoretically guaranteed to monotonically non-increase $J_{\mathrm{SL}}$ at each iteration.
We then develop a method that approximates this estimation approach and is computable in situations where $\bm{X}^c$ is not available.

\section{Proposed Method}
\label{sec:main_body}

In this section, we propose a method for estimating the exact values of CFs based upon the formalization presented in Section \ref{subsec:derivation_obj_func}.
When estimating the exact values of CFs using the derived objective function, these values are self-referenced during the optimization process.
In Section \ref{subsec:ideal_method}, under the hypothetical setting that the exact values of CFs can be self-referenced, we derive a confidence propagation method to estimate these values and theoretically validate its effectiveness.
This method involves the process that require self-referencing the exact values of CFs and the process that become computationally difficult when many CFs are considered. 
In Section \ref{subsec:approx_weight_matrix}, we approximate the former process using a method that can be computed without self-referencing the exact values of CFs. 
In Section \ref{subsec:approx_conf_prop}, we approximate the latter process with a method that remains practical even when many CFs are involved. 
Finally, in Section \ref{subsec:additional_components}, we discuss the relationships between multiple CFs and explain how to utilize the observed CF values.

\subsection{Retrieving the Exact values of CFs Under Hypothetical Setting}
\label{subsec:ideal_method}

In this section, we first develop a graph-based iterative method to retrieve the exact values of CFs under the hypothetical scenario where $\bm{X}^c$ can be self-referenced.
We then establish the theoretical validity of this approach.
The scenario that $\bm{X}^c$ can be self-referenced does not hold in the practical context addressed in this study.
Despite its impracticality, introducing this method allows us to outline guidelines for an estimation approach that remains computationally feasible even when $\bm{X}^c$ is unavailable.

Hereinafter, we illustrate the method using $n$ data points, denoted as $\{(\bar{\bm{x}}^{c}_i, \bm{x}^{o}_i, y_i)\}_{i\in [n]}$.
The conditional joint distribution of the exact values of CFs for each instance that we aim to recover is represented as $\{p_*(\bm{x}^{c}|\bm{x}^{o}_i)\}_{i \in [n]}$. 
We define the conditional joint confidence distribution (abbreviated as joint confidence) of the exact values of CFs for each instance as $\{q(\bm{x}^{c}|\bm{x}^{o}_i)\}_{i \in [n]}$. 
Here, $q(\bm{x}^{c}|\bm{x}^{o}_i)$ denotes the probability expressing the confidence that the exact value $\bm{x}^{c}_i$ of the CFs for the $i$-th instance is $\bm{x}^{c} \in \mathcal{X}^{c}$.

Using $q(\bm{x}^{c}|\bm{x}^{o}_i)$, we denote the following probability vector by $\bm{q}_{i} = [q(x^c_1=v_1,\dots,x^c_{F^c}=v_{F^c}|\bm{x}^{o}_i)]^{\top}_{v_j \in [|\mathcal{X}^{c}_j|], j \in [F^{c}]}$.
Defining such vector for all instances, we can denote the joint confidence matrix by $\bm{Q} = [\bm{q}_{1},\dots, \bm{q}_{n}]^{\top}$.

Our proposed method under the hypothetical scenario first sets each element of the initial joint confidence matrix $\bm{Q}^{(0)}=[q^{(0)}(\bm{x}^{c}|\bm{x}_i^{o})]_{n\times |\mathcal{X}^{c}|}$ based on observed values of CFs as follows:
\begin{align}
\begin{split}
    &q^{(0)}(\bm{x}^c|\bm{x}_i^{o}) = \begin{cases} \frac{1}{\prod^{F^c}_{j=1}(|\mathcal{X}^{c}_{j}|-1)} & \text{if } \bigwedge^{F^c}_{j=1}(x^c_j \neq \bar{x}^c_{ij}) \\ 0 & \text{otherwise} \end{cases}, \\
        & ~~~~~~~~~~~~~~~~~~~~ \forall \bm{x}^c \in \mathcal{X}^c,~~ \forall i \in [n]. 
\label{eq:init_conf_joint}
\end{split}
\end{align}
Next, $\bm{X}^c$ is retrieved through iterative confidence propagation, as follows:
Let $T$ denote the total number of iterations, and $\bm{Q}^{(t)}=[q^{(t)}(\bm{x}^c|\bm{x}_i^{o})]_{n\times  |\mathcal{X}^{c}|}$ denote the joint confidence matrix obtained in the $t$-th propagation for any $t \in [T]$.
The optimization of weight matrix $\bm{H}$ representing the similarity between instances and the confidence propagation, which are defined as shown below are executed iteratively:
\begin{align}
\begin{split}
    &\widehat{\bm{H}}^{(t)}_i \equiv \argmin_{\bm{H}_i}D_{\mathrm{KL}}\bigg( \sum^n_{i'=1}H_{ii'}q^{(t-1)}(\bm{X}^{c}|\bm{x}^{o}_{i'}) \bigg|\bigg|p_*(\bm{X}^{c}|\bm{x}^{o}_i) \bigg) \\
    &~~~ \text{s.t. } ~~ \sum^n_{i'=1}H_{ii'}=1 ~ \land ~ (H_{ii'} \ge 0 ~~\forall i' \in [n]), \\
    &~~~ \text{where }~~ \bm{H}_i = [H_{i1},...,H_{in}]^{\top}, ~~ \forall i \in [n],
\end{split}
\label{eq:weight_opt_ideal}
\end{align}
and
\begin{align}
    &\bm{Q}^{(t)} = \widehat{\bm{H}}^{(t)}\bm{Q}^{(t-1)}, ~ \text{where} ~ \widehat{\bm{H}}^{(t)} = [\widehat{\bm{H}}^{(t)}_1,...,\widehat{\bm{H}}^{(t)}_n]^{\top}.
    \label{eq:conf_prop_joint}
\end{align}

Eq. \eqref{eq:conf_prop_joint} can be expressed for any $i \in [n]$ as follows:
\begin{align}
    &q^{(t)}(\bm{x}^{c}|\bm{x}^{o}_i) = \sum^n_{i'=1}\widehat{H}^{(t)}_{ii'}q^{(t-1)}(\bm{x}^{c}|\bm{x}^{o}_{i'}), ~~ \forall \bm{x}^c \in \mathcal{X}^{c}.
    \label{eq:conf_prop_joint_detail}
\end{align}
From this representation, the following theorem holds for this retrieval method:
\begin{thm}
    For any $i \in [n]$ and any $t \in [T]$, the following holds:
     \begin{align}
     \begin{split}
         &D_{\mathrm{KL}}(p_*(\bm{X}^{c}|\bm{x}^{o}_i) || q^{(t)}(\bm{X}^{c}|\bm{x}^{o}_i)) \\
         &~~~~~~ \le D_{\mathrm{KL}}(p_*(\bm{X}^{c}|\bm{x}^{o}_i) || q^{(t-1)}(\bm{X}^{c}|\bm{x}^{o}_i) ).
     \end{split}
     \label{eq:interative_revision}
     \end{align}
\label{thm:iterative_revision}
\end{thm}
A proof is given in Appendix \ref{subsec:proof_iterative_revision} and intuitively interpreted as follows.
Because the confidence propagation is represented by Eq. \eqref{eq:conf_prop_joint_detail}, $q^{(t)}(\bm{x}^{c}|\bm{x}^{o}_i)$ can be interpreted as the predictive distribution resulting from the weighted average of the predictive distributions $\{q^{(t-1)}(\bm{x}^{c}|\bm{x}^{o}_{i'})\}_{i'\in [n]}$ of the exact values of CFs for each instance obtained in the $(t-1)$-th iteration.
Therefore, by determining the weight matrix $\widehat{\bm{H}}^{(t)}$ using Eq. \eqref{eq:weight_opt_ideal}, appropriate weighted averages are achieved in each propagation, leading to an improvement in the confidences for each instance.

From Theorem \ref{thm:iterative_revision}, the following corollary holds:
\begin{cor}
    For any $t \in [T]$, the following holds:
    \begin{align}
    \begin{split}
        &\frac{1}{n}\sum^{n}_{i=1}D_{\mathrm{KL}}(p_*(\bm{X}^{c}|\bm{x}^{o}_i) || q^{(t)}(\bm{X}^{c}|\bm{x}^{o}_i)) \\
        &~~~~~~ \le \frac{1}{n}\sum^{n}_{i=1} D_{\mathrm{KL}}(p_*(\bm{X}^{c}|\bm{x}^{o}_i) || q^{(t-1)}(\bm{X}^{c}|\bm{x}^{o}_i) ).
    \end{split}
    \end{align}
\label{cor:iterative_revision_sum}
\end{cor}

Corollary \ref{cor:iterative_revision_sum} implies that $J_{\mathrm{SL}}$ approximated using the empirical distribution of $n$ samples, monotonically non-increases at each iteration, ensuring the theoretical validity of the proposed approach.
However, because this method requires $\bm{X}^c$ to perform calculations using $\{p_*(\bm{x}^{c}| \bm{x}^{o}_i)\}_{i\in [n]}$ when optimizing $\{\bm{H}^{(t)}\}_{t \in [T]}$ through Eq. \eqref{eq:weight_opt_ideal}, it cannot be executed in practice, when $\bm{X}^c$ is not observed.
Additionally, the computational complexity of Eq.\eqref{eq:conf_prop_joint} increases exponentially with the number of CFs, hindering the execution of confidence propagation in situations involving many CFs.
To address the first issue, in Section \ref{subsec:approx_weight_matrix}, we set the smoothness assumption in SSL \cite{chapelle2006semi} to replace the optimization of $\{\bm{H}^{(t)}\}_{t \in [T]}$ using probability distributions on $\bm{X}^c$ with a computable optimization using instances on $\mathcal{X}^{o}$.
The approach to the second issue is described in Section \ref{subsec:approx_conf_prop}

\subsection{Practical Approximation of Weight Matrix Optimization}
\label{subsec:approx_weight_matrix}

To replace the optimization of $\{\bm{H}^{(t)}\}_{t \in [T]}$ using probability distributions on $\bm{X}^c$ with an optimization on $\mathcal{X}^{o}$, we apply the smoothness assumption in SSL \cite{chapelle2006semi}, which assumes that points that are similar in the input space are also similar in the output space.
We concretely describe this assumption as follows:
\begin{asm}[smoothness assumption]
    Let $d$ denote a distance on $\mathcal{X}^{o}$.
    For any instance $\bm{x}_i, i\in[n]$, there exists some $k_{i} \in \mathbb{N}^{+}$ such that the set of indices of the $k_{i}$-nearest neighbors with respect to distance $d$ on $\mathcal{X}^{o}$ of $\bm{x}_i$ is denoted as $B_{k_i}(\bm{x}^{o}_i)$.
    For any $k_{i}$-nearest neighbor points $\{\bm{x}_{i'_1}, \bm{x}_{i'_2} | i'_1, i'_2 \in B_{k_i}(\bm{x}^{o}_i)\}$, the following holds:
    \begin{align}
    \begin{split}
        &d(\bm{x}^{o}_i, \bm{x}^{o}_{i'_1}) \le d(\bm{x}^{o}_i, \bm{x}^{o}_{i'_2}) \Rightarrow \\
        & D_{\mathrm{KL}}(p_*(\bm{X}^{c}|\bm{x}^{o}_{i'_1})||p_*(\bm{X}^{c}|\bm{x}^{o}_{i})) \\
        &~~~~~~~~~~~~~~~~~~~~ \le D_{\mathrm{KL}}(p_*(\bm{X}^{c}|\bm{x}^{o}_{i'_2})||p_*(\bm{X}^{c}|\bm{x}^{o}_{i})).
    \end{split}
    \label{eq:smoothness}
    \end{align}
    \label{asm:smoothness}
\end{asm}

Under this assumption, the positional relationships between instances in $\mathcal{X}^{o}$ correspond to the positional relationships between probability distributions on $\bm{X}^{c}$.
When the confidence matrix consistently improves at each propagation step, under Assumption \ref{asm:smoothness}, the equation that replaces $p_*(\bm{X}^{c}|\bm{x}^{o}_{i'_1})$ with $q^{(t-1)}(\bm{X}^{c}|\bm{x}^{o}_{i'_1})$ and $p_*(\bm{X}^{c}|\bm{x}^{o}_{i'_2})$ with $q^{(t-1)}(\bm{X}^{c}|\bm{x}^{o}_{i'_2})$ in Eq. \eqref{eq:smoothness} can be expected to hold for any $t \in [T]$.
Thus, when optimizing the weight matrix $\{\bm{H}^{(t)}\}_{t \in [T]}$, for any $i \in [n]$, we replace the problem of approximating $p_*(\bm{x}^{c}|\bm{x}^{o}_i)$ as a linear mixture of $\{q^{(t-1)}(\bm{x}^{c}|\bm{x}^{o}_{i'})\}_{i'\in [n]}$, expressed by Eq. \eqref{eq:weight_opt_ideal}, with the problem of approximating $\bm{x}^{o}_i$ as a linear mixture of $\{\bm{x}^{o}_{i'}\}_{i' \in B_{k_i}(\bm{x}^{o}_i)}$. 
In this process, for qualitative features included in $\bm{x}^{o}$ that take three or more values, distances must be computed in the same manner as quantitative features. 
Therefore, during the optimization, we use $\tilde{\bm{x}}^{o}$ obtained by replacing these features in $\bm{x}^{o}$ with their OneHot representations.
Let $\widetilde{\mathcal{X}}^{o}$ be defined as the space from which $\tilde{\bm{x}}^{o}_i$ can take values.
Specifically, we designate the Euclidean distance as a distance $d$ on $\widetilde{\mathcal{X}}^{o}$ and solve the following alternative optimization problem:
\begin{align}
\begin{split}
    &\widehat{\bm{H}}_{i} \equiv \argmin_{\bm{H}_i} \bigg\| \tilde{\bm{x}}^{o}_i - \sum_{i' \in B_{k_i}(\tilde{\bm{x}}^{o}_i)}H_{ii'}\tilde{\bm{x}}^{o}_{i'} \bigg\|_{2}  \\
    & \text{s.t.} ~ \sum_{i' \in B_{k_i}(\tilde{\bm{x}}^{o}_i)}H_{ii'}=1 ~ \land ~  (H_{ii'} \ge 0, ~ \forall i' \in B_{k_i}(\tilde{\bm{x}}^{o}_i))  \\
    & \text{where }~~ \bm{H}_i = [H_{i1},...,H_{in}]^{\top}, ~~ \forall i \in [n].
    \label{eq:approx_opt}
\end{split}
\end{align}

The optimization described in Eq. \eqref{eq:weight_opt_ideal} does not necessarily maintain $q^{(t-1)}$ unchanged in each iteration, but $\{\tilde{\bm{x}}^{o}_i\}_{i\in [n]}$ remains constant across iterations.
Therefore, if Eq. \eqref{eq:weight_opt_ideal} is substituted with Eq. \eqref{eq:approx_opt} in every iteration, the weight matrix $\widehat{\bm{H}}=[\widehat{\bm{H}}_1,\dots, \widehat{\bm{H}}_n]^{\top}$ obtained at each iteration will be the same.
In other words, using this alternative approach, all $\{\bm{H}^{(t)}\}_{t \in [T]}$ obtained from Eq. \eqref{eq:weight_opt_ideal} are approximated with a single weight matrix $\widehat{\bm{H}}$ obtained from Eq. \eqref{eq:approx_opt}.
By replacing the optimization of Eq. \eqref{eq:weight_opt_ideal} for any iteration $t \in [T]$ with the single optimization of Eq. \eqref{eq:approx_opt}, this approach significantly reduces computational complexity.
Therefore, this approximation is not only reasonable under Assumption \ref{asm:smoothness}, but also beneficial in terms of computational complexity.

The effectiveness of this optimization method depends on the scaling of the features used.
To ensure each feature effects equally, the following preprocessing methods are applied: 
i) quantitative features are scaled to fit within the range $[0, 1]$; 
ii) binary features are set to take values of $0$ or $1$; and 
iii) features represented by OneHot vectors are scaled by multiplying them with $1/\sqrt{|\mathcal{X}^{o}_{j}|}$, where $j$ is the index of each feature.
This scaling adjustment ensures that the squared norm of the difference between OneHot vectors to be on the same scale as the squared difference between values of quantitative features.

When performing optimization using Eq. \eqref{eq:approx_opt}, we must determine $\{k_i\}_{i\in [n]}$.
We substituted them with a single hyperparameter $k \in \mathbb{N}^{+}$ for the following reasons:
Eq. \eqref{eq:approx_opt} is intuitively expected to assign smaller weights to more distant neighboring points during the optimization process.
Therefore, even if $k$ is set to be bigger than $k_i$ satisfying assumption \ref{asm:smoothness}, it is considered that the influence of the extra neighboring points can be suppressed.
Thus, assigning a sufficiently large value to $k$ is deemed adequate.

\subsection{Practical Approximation of Confidence Propagation}
\label{subsec:approx_conf_prop}

While the optimization of the weight matrix for confidence propagation has become feasible with Eq. \eqref{eq:approx_opt}, subsequent confidence propagation using Eq. \eqref{eq:conf_prop_joint} faces two significant challenges due to the dimension $|\mathcal{X}^{c}|$ of the probability vector representing the joint confidence distribution, which becomes excessively large in realistic scenarios. 
The first challenge arises from the computational complexity of Eq. \eqref{eq:conf_prop_joint}, which scales as $O(Tn^2|\mathcal{X}^{c}|)$.
This leads to prohibitively large computational requirements for executing Eq. \eqref{eq:conf_prop_joint}. 
The second challenge stems from using the confidences obtained from Eq. \eqref{eq:conf_prop_joint} as inputs to the label prediction model $p_{\bm{\theta}}$, resulting in excessively large input dimensions for $p_{\bm{\theta}}$. 
This exacerbates the curse of dimensionality.
Given that $|\mathcal{X}^{c}| =  \prod_{j\in [F^{c}]}|\mathcal{X}^{c}_j|$, $|\mathcal{X}^{c}|$ exponentially increases with the number $|F^{c}|$ of CFs. 
For instance, in the experiments detailed in Section \ref{sec:experiments}, datasets with 5 and 7 CFs yield $|\mathcal{X}^{c}|$ values of $1728$ and $19051200$, respectively. 
Thus, in practical scenarios, executing confidence propagation and effectively utilizing its results becomes challenging.

Regarding the second challenge, two approaches can be considered: 
{\it argmax}, which uses the OneHot representation of $\hat{\bm{x}}^{c}_i = \argmax_{\bm{x}^{c}}q^{(T)}(\bm{x}^{c}|\bm{x}^{o}_i)$ as input, and 
{\it marginalization}, which calculates the confidence vector for each $j$-th CF using $\sum_{\bm{x}^c_{\setminus j} \in \mathcal{X}^{c}_{\setminus j}}q^{(T)}(\bm{x}^c_{\setminus j}, x^{c}_j|\bm{x}^{o}_i)$. 
Here, $\setminus j$ represents the set of indices of CFs excluding the $j$-th one.
Both approaches can reduce the input dimension concerning CF for $p_{\bm{\theta}}$ to $\sum_{j \in [F^{c}]}|\mathcal{X}^{c}_j|$.
The drawbacks of {\it argmax} are the loss of confidence information regarding the exact value of CFs and the estimation errors that can occur due to selecting a single estimated value. 
These issues reduce both input information and the interpretability of $p_{\bm{\theta}}$'s output.
On the other hand, {\it marginalization} loses the information about the relationships between multiple CFs by marginalizing joint confidences but retains the confidence information for each CF under $X^{c}_1 \indep \cdots \indep X^{c}_{F^c} |\bm{X}^{o}$.
Therefore, to address the second challenge, {\it marginalization} is considered more effective.

In addressing the second challenge, rather than using the joint confidence distributions, it is sufficient in practice to obtain the conditional marginal confidence distributions (abbreviated as marginal confidences) $\{\sum_{\bm{x}^c_{\setminus j} \in \mathcal{X}^{c}_{\setminus j}}q^{(T)}(\bm{x}^c_{\setminus j}, x^{c}_j | \bm{x}^{o}_i)\}_{i \in [n], j \in [F^c]}$.
Therefore, by considering the propagation of marginal confidences for each $j$-th CF, ($j \in [F^{c}]$), we aim to tackle the first challenge. 
First, we set the initial value of marginal confidence $\{\bm{Q}^{(0)}_{(j)} = [q^{(0)}(x^c_j|\bm{x}^{o}_i)]_{n\times |\mathcal{X}^{c}_j|}\}_{j\in F^c}$ as follows:
\begin{align}
\begin{split}
    &q^{(0)}(x^c_j|\bm{x}^{o}_i) = \begin{cases}
        \frac{1}{|\mathcal{X}^c_j|-1} & \text{if } x^{c}_j \neq \bar{x}^c_{ij} \\
        0 & \text{if } x^{c}_j = \bar{x}^c_{ij}
    \end{cases},  \\
    &~~~~~~~~~~~~~~~~~~~~~ \forall i \in [n], \forall j \in [F^{c}].
\end{split}
\label{eq:init_conf_marginal}
\end{align}
For any $j \in [F^{c}]$ and $t \in \{0,1,\dots,T\}$, we define the marginal confidence matrix in the $t$-th propagation of the $j$-th CF as  $\bm{Q}^{(t)}_{(j)} \equiv [q^{(t)}(x^{c}_j | \bm{x}^{o}_i)]_{n \times |\mathcal{X}^c_j|}$.
Confidence propagation using marginal confidence is defined by the following equation:
\begin{align}
    \bm{Q}^{(t)}_{(j)} = \widehat{\bm{H}} \bm{Q}^{(t-1)}_{(j)}, \forall t \in [T], \forall j \in [F^{c}].
\label{eq:conf_prop_marginal}
\end{align}  
Describing Eq. \eqref{eq:conf_prop_marginal} element-wise, gives:
\begin{align}
   &q^{(t)}(x^{c}_j|\bm{x}^{o}_i) = \sum^n_{i'=1,i'\neq i}\hat{H}_{ii'}q^{(t-1)}(x^{c}_j|\bm{x}^{o}_{i'}), ~~ \forall i \in [n]. 
\label{eq:conf_prop_marginal_detail}
\end{align} 
Because the marginal confidence for each CF is represented by an $|\mathcal{X}^{c}_j|$-dimensional confidence vector, the computational complexity of confidence propagation for each CF is $O(Tn^2|\mathcal{X}^{c}_j|)$. 
Therefore, by substituting the propagation of marginal confidence for the propagation of joint confidence, the computational complexity reduces from $O(Tn^2\prod_{j\in [F^{c}]}|\mathcal{X}^{c}_j|)$ to $O(Tn^2\sum_{j \in [F^{c}]}|\mathcal{X}^{c}_j|)$.
In the experiments discussed in Section \ref{sec:experiments}, wherein two datasets were used, the values of $\prod_{j\in [F^{c}]}|\mathcal{X}^{c}_j|$ were $1728$ and $19051200$, whereas those of $\sum_{j \in [F^{c}]}|\mathcal{X}^{c}_j|$ were $26$ and $100$, indicating a significant reduction in computational complexity.
Thus, confidence propagation using marginal confidence is considered a practical solution to both challenges.

Moreover, confidence propagation using marginal confidence is not only practically executable but is also theoretically validated by the following theorem.
\begin{thm}
    For any $i \in [n], j \in [F^{c}], t \in \{0,1,\dots,T\}$, 
    and with the joint confidence matrix $\bm{Q}^{(t)}=[q^{(t)}(\bm{x}^c|\bm{x}_i^{o})]_{n\times  |\mathcal{X}^{c}|}$ obtained from Eq. \eqref{eq:init_conf_joint} or Eq. \eqref{eq:conf_prop_joint}, and the marginal confidence matrix $\bm{Q}^{(t)}_{(j)}=[q^{(t)}(x^{c}_j | \bm{x}^{o}_i)]_{n \times |\mathcal{X}^c_j|}$ obtained from Eq. \eqref{eq:init_conf_marginal} or Eq. \eqref{eq:conf_prop_marginal}, the following holds:
    \begin{align}
        q^{(t)}(x^{c}_j | \bm{x}^{o}_i) =  \sum_{\bm{x}^c_{\setminus j} \in \mathcal{X}^{c}_{\setminus j}}q^{(t)}(\bm{x}^c_{\setminus j}, x^{c}_j|\bm{x}^{o}_i).
        \label{eq:joint_marginal_equality}
    \end{align}
\label{thm:joint_marginal_equality}
\end{thm}
A proof is given in Appendix \ref{subsec:proof_joint_marginal_equality}.
Theorem \ref{thm:joint_marginal_equality} shows that the marginal confidence obtained by propagating joint confidence represented (Eq. \eqref{eq:conf_prop_joint}) is equivalent to the marginal confidence obtained by propagating marginal confidence (Eq. \eqref{eq:conf_prop_marginal}). 
Therefore, using only marginal confidence as input to the label prediction model $p_{\bm{\theta}}$, using Eq. \eqref{eq:conf_prop_marginal} instead of Eq. \eqref{eq:conf_prop_joint} to perform confidence propagation is theoretically justified.
Based on the above, in our proposed method, after optimizing the weight matrix with Eq. \eqref{eq:approx_opt}, we estimate the confidence for the exact values of CFs by performing confidence  propagation, as expressed by Eq. \eqref{eq:conf_prop_marginal}.
Additionally, the single estimated exact value for each CF is determined by the value with the highest confidence.

\begin{figure}[t]
\begin{algorithm}[H]
    \caption{Proposed Method}
    \label{alg:propose}
    \begin{algorithmic}
    \Inputs{
        $S$: the sample set $\{(\bar{\bm{x}}^c_i, \bm{x}^{o}_i, y_i)\}_{i\in [n]}$ \\
        $k$: the number of neighbors \\
        $\gamma$: the weight for distance calculations including CFs' estimated exact values in $[0,1]$ \\
        $T$: the number of iterations \\   
    }
    \Outputs{
        $\big\{\bm{Q}^{(T)}_{(j)}\big\}_{j \in [F^{c}]}$: confidence matrices of the CFs' exact values \\
        $\{\hat{\bm{x}}^c_i\}_{i\in [n]}$: Estimated values for the CFs' exact values
    }
    \Statex
    \Process{
        Initialize Confidence Matrix $\big\{\bm{Q}^{(0)}_{(j)}\big\}_{j \in [F^{c}]}$ by Eq.\eqref{eq:init_conf_marginal} \\ 
        \For{$i\in [n]$}
            \State Identify $B_k(\bm{x}_i)$ using $\{\bm{x}^{o}_i\}_{i\in [n]}$
            \State Determine the weight matrix $\widehat{\bm{H}}_i$ according to Eq.\eqref{eq:approx_opt}
        \EndFor
        \State Set $\widehat{\bm{H}} = [\widehat{\bm{H}}_1,\dots,\widehat{\bm{H}}_n]^{\top}$
        \For{$t \in [T]$}
            \State Set $\big\{\bm{Q}^{(t)}_{(j)}\big\}_{j \in [F^{c}]}$ according to Eq.\eqref{eq:conf_prop_marginal}
            \State Update $\big\{\bm{Q}^{(t)}_{(j)}\big\}_{j \in [F^{c}]}$ via Eq.\eqref{eq:correct_own_comp}
        \EndFor
        \For{$i \in [n]$}
            \State Identify $B_k(\bm{x}_i)$ 
            \State \hspace{\algorithmicindent} using $\{\bm{x}^{o}_{i'}\}_{i'\in [n]}$ and $\big\{\bm{Q}^{(T)}_{(j)}\big\}_{j \in [F^{c}]}$ with $\gamma$
            \State Determine the weight matrix $\widehat{\bm{H}}_i$ according to Eq.\eqref{eq:approx_opt}
            \State \hspace{\algorithmicindent} using $\{\bm{x}^{o}_{i'}\}_{i'\in [n]}$ and $\big\{\bm{Q}^{(T)}_{(j)}\big\}_{j \in [F^{c}]}$ 
        \EndFor
        \State Set $\widehat{\bm{H}} =[\widehat{\bm{H}}_1,\dots,\widehat{\bm{H}}_n]^{\top}$
        \For{$t \in [T]$}
            \State Set $\big\{\bm{Q}^{(t)}_{(j)}\big\}_{j \in [F^{c}]}$ according to Eq.\eqref{eq:conf_prop_marginal}
            \State Update $\big\{\bm{Q}^{(t)}_{(j)}\big\}_{j \in [F^{c}]}$ via Eq.\eqref{eq:correct_own_comp}
        \EndFor
        \State Determine $\{\hat{\bm{x}}^c_i\}_{i\in [n]}$ from $\{\bm{Q}^{(T)}_{(j)}\}_{j \in [F^c]}$
    }
    
    \end{algorithmic}
\end{algorithm}
\end{figure}

\subsection{Additional Tactics}
\label{subsec:additional_components}

The above approximation method does not consider relationships between multiple CFs.
This is because, when estimating the exact values of a particular CF, the method utilizes only observed values of OFs and the CF itself, neglecting information from observed values of other CFs.
Therefore, the approximation method derived from Eq. \eqref{eq:approx_opt} is performed again using $\{\hat{\bm{x}}_i = (\hat{\bm{x}}^c_i, \bm{x}^{o}_i)\}_{i\in [n]}$, where $\hat{\bm{x}}^c_i$ denotes the estimated exact values of CFs.
This method allows for the use of information from other CFs to estimate the exact value of each CF through their estimated exact values. 
In other words, this approach enables estimation that considers the interdependencies among multiple CFs.

However, this method encounters two primary issues.
Firstly, $\hat{\bm{x}}^{c}_i$ representing a single estimated value, fails to account for the inherent uncertainty in the estimation outcomes.
To address this, we replace the OneHot vector representation of $\hat{\bm{x}}^{c}_i$ used in solving Eq. \eqref{eq:approx_opt}, with the estimated confidences $\{q^{(T)}(x^{c}_j|\bm{x}^{o}_i)\}_{j \in [F^c]}$, thereby incorporating uncertainty into the estimation results.
Secondly, treating the estimated value $\hat{\bm{x}}^{c}_i$ and the observed value $\bm{x}^{o}_i$ as equal disregards the estimation error of $\hat{\bm{x}}^{c}_i$.
When $\{\hat{\bm{x}}^{c}_i\}_{i\in [n]}$ contain numerous errors, the weight matrix $\bm{H}$, optimized using $\{\hat{\bm{x}}_i\}_{i\in [n]}$ in Eq. \eqref{eq:approx_opt}, may capture similarities between samples on $\mathcal{X}$ that diverge from their actual similarities. 
Consequently, this could lead to inappropriate confidence propagation.
To mitigate the impact of estimation errors, we introduce a hyperparameter $\gamma$.
By weighting $\{\hat{x}^c_{ij}\}_{i\in [n], j\in [F^{c}]}$ with $\gamma$, we regulate the impact of these features in the optimization of the weight matrix.
Specifically, employing Euclidean distance, each confidence vector in $\{\hat{x}^c_{ij}\}_{i\in [n]}$ is multiplied by $\sqrt{\gamma}$ to ensure that their norm is scaled by a factor of $\gamma$.

\begin{table}[t]
    \centering
    \caption{Outline of datasets.}
    \begin{tabular}{c|c|c}
        \hline
         dataset   & Bank & Adult \\ \hline \hline
         data size & 45211 & 48842 \\ \hline 
         the number of binary input features & 3 & 1 \\ \hline
         the number of categorical input features & 5 & 7 \\ \hline
         the number of numerical input features & 7 & 6 \\ \hline
         type of the target feature & binary & binary \\ \hline
    \end{tabular}
    \label{tab:outline_datasets}
\end{table}

This approximation method does not directly utilize the CF value of an instance to update its confidence.
Instead, the CF value for a specific instance indirectly influences the confidence of its exact value through iterative confidence propagation.
Consequently, during the $t$-th propagation, ($t \in [T]$), for any $i\in [n]$ and $j\in [F^{c}]$, there is a possibility that $q^{(t)}(\bar{x}^{c}_{ij}|\bm{x}^{o}_i)$, which should be $0$, may become positive.
Therefore, following the confidence propagation for the $t$-th iteration, ($t \in [T]$), $\bm{Q}^{(t)}_{(j)} $ is updated using the following operation:
\begin{align}
    \bm{Q}^{(t)}_{(j)}
    &\equiv \text{Normalize}_{\mathrm{r}}\big(\bm{Q}^{(t)}_{(j)} \odot \bm{Q}^{(0)}_{(j)}\big), ~~ \forall j \in [F^{c}].
    \label{eq:correct_own_comp}
\end{align}
Here, $\text{Normalize}_{\mathrm{r}}(\cdot)$ is a function that normalizes matrices row-wise, and the operator $\odot$ denotes the Hadamard product.
This process distributes the value of $q^{(t)}_{(j)}(\bar{x}^{c}_{ij} | \bm{x}^{o}_i)$ evenly among each $\{q^{(t)}_{(j)}(x^{c}_{j} | \bm{x}^{o}_i)\}_{x^{c}_{j} \in \mathcal{X}^{c}_j \setminus \{\bar{x}^{c}_{ij}\}}$ for any $i$-th sample, ($i \in [n]$).
When the probability of the observed value of a CF is $0$ in the distribution of the exact value of the CF, this operation is expected to decrease $J_{\mathrm{SL}}$.

Our proposed method is summarized by Algorithm \ref{alg:propose}.
The main differences between our proposed method and existing graph-based SSL and PLL methods \cite{zhu2002learning,zhang2015solving}, apart from being derived from the formulation of CFL, are twofold:
Firstly, our proposed method considers the relationships between multiple CFs, whereas existing SSL or PLL methods do not account for these relationships.
Secondly, the treatment of initial confidence $\{\bm{Q}^{(0)}_{(j)}\}_{j\in [F^c]}$ differs significantly.
In existing methods applied to CFL, for any $j \in [F^c]$ and $t\in\{2,\dots,T\}$, $\bm{Q}^{(t)}_{(j)}$ is determined by a linear combination of $\widehat{\bm{H}}\bm{Q}^{(t-1)}_{(j)}$ and $\bm{Q}^{(0)}_{(j)}$, thus incorporating initial confidence values to some extent in each propagation \cite{zhu2002learning,zhang2015solving}.
This approach tends to reduce the confidences of values that are known not to be exact values at initialization.
In contrast, our proposed method reflects initial confidences using Eq. \eqref{eq:correct_own_comp}, ensuring that elements with an initial confidence of $0$ are corrected to $0$ in each obtained confidence vector.
Therefore, our method facilitates a more effective use of information from initial values.

\begin{figure*}[t]
    \centering
    \includegraphics[width=0.9\linewidth]{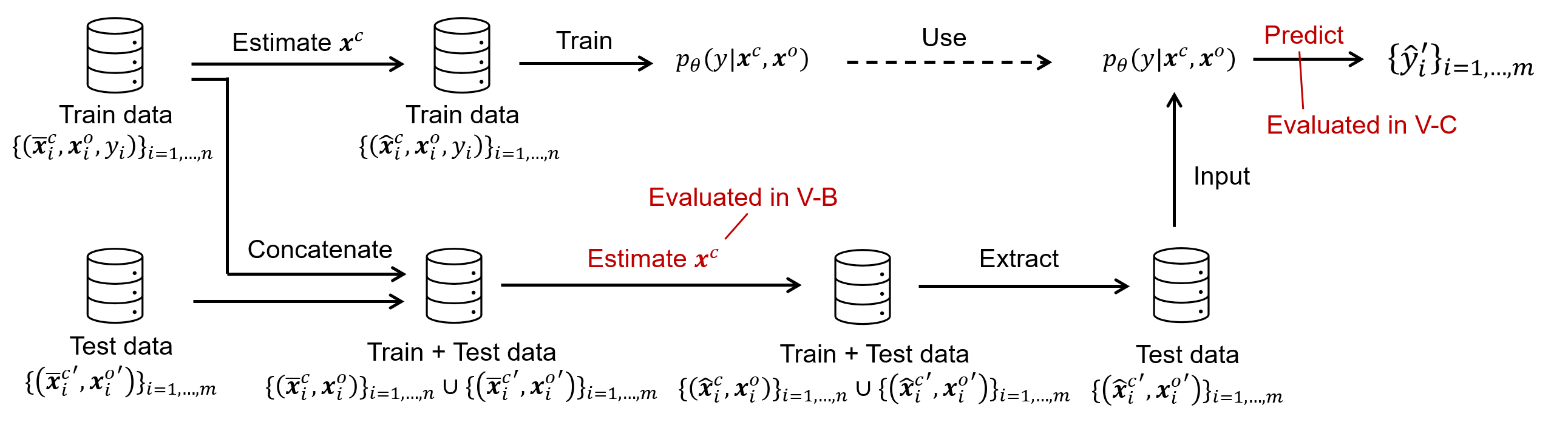}
    \caption{The process of evaluating our proposed method.}
    \label{fig:experimental_process}
\end{figure*}

\section{Numerical experiments}
\label{sec:experiments}

In this section, we evaluate our proposed method using real-world datasets.
In Section \ref{subsec:exp_settings}, we describe the evaluation datasets and experimental settings.
In Section \ref{subsec:exp_estimate_CF}, we evaluate the performance of our proposed method in estimating the exact values of CFs.
In Section \ref{subsec:exp_predict_target}, we evaluate the accuracy of output labels predicted using the estimated exact values of CFs obtained through our proposed method.
The implementation of this experiment is available at \url{https://github.com/KOHsEMP/learning_from_cf}.

\subsection{Experimental Settings}
\label{subsec:exp_settings}

For evaluation, we used the Bank Marketing (Bank) and Adult datasets provided by the UCI repository \cite{Dua:2019}.
These datasets contain features related to personal information, such as occupation, race, and education level, allowing us to simulate scenarios where precise personal information is unavailable due to privacy concerns.
The specifications of the two datasets are summarized in Table \ref{tab:outline_datasets}.
Here, categorical input features are qualitative and non-binary.
In our experiments, all categorical input features were converted to CFs.
During this conversion, values other than the true value were randomly assigned as CF values.

Figure \ref{fig:experimental_process} illustrates the execution flow of our proposed method in this experiment.
We assumed that the training data was available during the inference of the test data, which made IPAL, a comparison method, executable.
We treated half of the data as training data for the label prediction model $p_{\bm{\theta}}$, whereas the remaining half served as test data.

We compared the proposed method ({\it proposed}), the direct utilization of CFs' observed values ({\it comp}), and IPAL \cite{zhang2015solving} which is based on confidence propagation in the PLL, as estimation methods for the exact values of CFs.
The {\it comp} method, adopted as the baseline, uses the initial confidence $\{\bm{Q}^{(0)}_{(j)}\}_{j \in [F^c]}$, obtained from Eq. \eqref{eq:init_conf_marginal}, as the estimation result for the CFs' exact values.
IPAL is a method that predicts output labels using confidence propagation and was selected for comparison with the proposed method because it can also estimate CFs' exact values.
IPAL performs confidence propagation to predict output labels in the training data.
Its hyperparameters include $T$, $k$, and a balancing coefficient $\alpha' \in (0,1)$.
Confidence propagation using training data is performed as $\bm{Q}^{(t)}_{(j)} = \alpha'\hat{\bm{H}}\bm{Q}^{(t-1)}_{(j)} + (1-\alpha')\bm{Q}^{(0)}_{(j)},  \forall t \in [T], j \in [F^{c}]$.

The $T$ and $k$ in IPAL have the same meaning as those in the proposed method.
On the other hand, when predicting output labels for the test data, IPAL does not perform confidence propagation; instead, it determines a single predicted value based on the predicted labels of the training data \cite{zhang2015solving}.
Consequently, when applying IPAL to estimate CFs' exact values, the confidence in the CFs' exact value cannot be used as input to the label prediction model.

\subsection{Evaluation of Estimation Quality for CF}
\label{subsec:exp_estimate_CF}

\begin{table*}[t]
    \caption{Estimation Performance for each CF (Bank)}
    \centering
    \begin{tabular}{l|l|l|l|l|l} 
    \hline
    Feature & Method & Acc & F1 & CE & SE  \\ 
    \hline \hline 
    \multirow{3}{*}{job (12)} & comp  & 0.0903  ±0.0014 & 0.0744  ±0.0010 & 2.3979  ±0.0000 & 2.3979  ±0.0000 \\ 
                                   & IPAL  & 0.1831  ±0.0092 & 0.1276  ±0.0043 & 2.4649  ±0.0007 & 2.4837  ±0.0000 \\ 
                                   & proposed  & {\bf  0.2543  ±0.0093} & {\bf  0.1393  ±0.0041} & {\bf  2.2253  ±0.0130} & {\bf  2.0430  ±0.0366} \\ 
    \hline 
    \multirow{3}{*}{marital (3)} & comp  & 0.4997  ±0.0024 & 0.4538  ±0.0022 & 0.6931  ±0.0000 & 0.6931  ±0.0000 \\ 
                                   & IPAL  & 0.7183  ±0.0012 & 0.4963  ±0.0025 & 0.9387  ±0.0007 & 1.0762  ±0.0002 \\ 
                                   & proposed  & {\bf  0.7606  ±0.0014} & {\bf  0.5720  ±0.0060} & {\bf  0.6405  ±0.0100} & {\bf  0.3162  ±0.0079} \\ 
    \hline 
    \multirow{3}{*}{education (4)} & comp  & 0.3328  ±0.0020 & 0.2863  ±0.0017 & 1.0986  ±0.0000 & 1.0986  ±0.0000 \\ 
                                   & IPAL  & 0.5804  ±0.0032 & 0.3714  ±0.0061 & 1.2948  ±0.0006 & 1.3785  ±0.0001 \\ 
                                   & proposed  & {\bf  0.5989  ±0.0019} & {\bf  0.3792  ±0.0067} & {\bf  1.0625  ±0.0208} & {\bf  0.5705  ±0.0084} \\ 
    \hline 
    \multirow{3}{*}{contact (3)} & comp  & 0.5013  ±0.0032 & 0.4268  ±0.0031 & 0.6931  ±0.0000 & 0.6931  ±0.0000 \\ 
                                   & IPAL  & 0.8819  ±0.0007 & 0.6393  ±0.0030 & 0.8097  ±0.0013 & 1.0549  ±0.0004 \\ 
                                   & proposed  & {\bf  0.9112  ±0.0016} & {\bf  0.7442  ±0.0083} & {\bf  0.3862  ±0.0032} & {\bf  0.1076  ±0.0021} \\ 
    \hline 
    \multirow{3}{*}{poutcome (4)} & comp  & 0.3337  ±0.0018 & 0.2199  ±0.0014 & 1.0986  ±0.0000 & 1.0986  ±0.0000 \\ 
                                   & IPAL  & 0.9193  ±0.0017 & 0.5709  ±0.0143 & 1.1357  ±0.0004 & 1.3672  ±0.0001 \\ 
                                   & proposed  & {\bf  0.9234  ±0.0011} & {\bf  0.5941  ±0.0118} & {\bf  0.2829  ±0.0084} & {\bf  0.0946  ±0.0030} \\ 
    \hline 
    \end{tabular}
    \label{tab:cf_score_bank}
\end{table*}

\begin{table*}[t]
    \caption{Estimation Performance for each CF (Adult)}
    \centering
    \begin{tabular}{l|l|l|l|l|l} 
    \hline
    Feature & Method & Acc & F1 & CE & SE  \\ 
    \hline \hline 
    \multirow{3}{*}{workclass (9)} & comp  & 0.1250  ±0.0013 & 0.0750  ±0.0009 & 2.0794  ±0.0000 & 2.0794  ±0.0000 \\ 
                                   & IPAL  & 0.6469  ±0.0041 & {\bf  0.1409  ±0.0033} & 2.1336  ±0.0006 & 2.1953  ±0.0000 \\ 
                                   & proposed  & {\bf  0.6896  ±0.0022} & 0.1342  ±0.0047 & {\bf  1.4867  ±0.0133} & {\bf  0.4911  ±0.0222} \\ 
    \hline 
    \multirow{3}{*}{education (16)} & comp  & 0.0648  ±0.0009 & 0.0434  ±0.0006 & {\bf  2.8078  ±0.0000} & 2.7215  ±0.0000 \\ 
                                   & IPAL  & {\bf  0.1608  ±0.0029} & {\bf  0.0643  ±0.0030} & 2.8351  ±0.0003 & 2.8282  ±0.0000 \\ 
                                   & proposed  & 0.0176  ±0.0018 & 0.0061  ±0.0022 & 3.3628  ±0.0328 & {\bf  1.7085  ±0.0147} \\ 
    \hline 
    \multirow{3}{*}{marital (7)} & comp  & 0.1649  ±0.0023 & 0.1129  ±0.0016 & 1.7918  ±0.0000 & 1.7918  ±0.0000 \\ 
                                   & IPAL  & 0.6469  ±0.0020 & 0.3226  ±0.0057 & 1.8637  ±0.0006 & 1.9429  ±0.0000 \\ 
                                   & proposed  & {\bf  0.6776  ±0.0031} & {\bf  0.3390  ±0.0027} & {\bf  1.2602  ±0.0262} & {\bf  0.4593  ±0.0162} \\ 
    \hline 
    \multirow{3}{*}{occupation (15)} & comp  & 0.0725  ±0.0009 & 0.0626  ±0.0010 & {\bf  2.6391  ±0.0000} & 2.6391  ±0.0000 \\ 
                                   & IPAL  & 0.1109  ±0.0049 & 0.0858  ±0.0050 & 2.6954  ±0.0002 & 2.7072  ±0.0000 \\ 
                                   & proposed  & {\bf  0.1423  ±0.0083} & {\bf  0.0986  ±0.0028} & 2.7329  ±0.0196 & {\bf  2.2846  ±0.0103} \\ 
    \hline 
    \multirow{3}{*}{relationship (6)} & comp  & 0.2004  ±0.0031 & 0.1708  ±0.0024 & 1.6094  ±0.0000 & 1.6094  ±0.0000 \\ 
                                   & IPAL  & 0.5903  ±0.0043 & 0.3932  ±0.0042 & 1.7100  ±0.0014 & 1.7880  ±0.0001 \\ 
                                   & proposed  & {\bf  0.6174  ±0.0022} & {\bf  0.3971  ±0.0063} & {\bf  1.3875  ±0.0233} & {\bf  0.5231  ±0.0206} \\ 
    \hline 
    \multirow{3}{*}{race (5)} & comp  & 0.2485  ±0.0026 & 0.1336  ±0.0015 & 1.3863  ±0.0000 & 1.3863  ±0.0000 \\ 
                                   & IPAL  & 0.8579  ±0.0010 & 0.2219  ±0.0053 & 1.4469  ±0.0005 & 1.6015  ±0.0001 \\ 
                                   & proposed  & {\bf  0.8730  ±0.0018} & {\bf  0.2701  ±0.0050} & {\bf  0.9034  ±0.0270} & {\bf  0.0996  ±0.0063} \\ 
    \hline 
    \multirow{3}{*}{native-country (42)} & comp  & 0.0245  ±0.0006 & 0.0049  ±0.0006 & 3.7136  ±0.0000 & 3.7136  ±0.0000 \\ 
                                   & IPAL  & 0.6726  ±0.0158 & 0.0226  ±0.0007 & 3.7180  ±0.0003 & 3.7374  ±0.0000 \\ 
                                   & proposed  & {\bf  0.8504  ±0.0062} & {\bf  0.0240  ±0.0012} & {\bf  2.6909  ±0.0252} & {\bf  3.5823  ±0.0024} \\ 
    \hline 
    \end{tabular}
    \label{tab:cf_score_adult}
\end{table*}

In this section, we evaluated the performance of our proposed method in estimating the CFs' exact values.
As depicted in Figure \ref{fig:experimental_process}, our proposed method utilized the entire dataset, combining both training and test data, for comprehensive evaluation.
The evaluation metrics employed include accuracy score (Acc), macro-F1 score (F1), cross entropy (CE), and Shannon entropy (SE).
A low SE coupled with high Acc and F1 indicates effective reduction of confidences for values other than the exact values.
The hyperparameters chosen for our proposed method were set as follows: $T=100, k=20, \gamma=0.25$.
For IPAL, the hyperparameters were set as: $T=100, k=20, \alpha'=0.9$.

Table \ref{tab:cf_score_bank} presents the evaluation results for the Bank dataset, whereas Table \ref{tab:cf_score_adult} presents those for the Adult dataset.
The headers of these tables correspond to the names of CFs, with the number of unique values each feature can take enclosed in parentheses.
The scores in the tables represent the averages and the standard deviations obtained from five trials.
Acc and F1 are calculated based on a single estimate, whereas CE and SE reflect the estimated confidence. 
Therefore, the following correspondences were taken for each comparison method: 
{\it proposed} used a single estimate determined by the value with the highest estimated confidence and used the estimated confidence directly.
{\it comp} used a value randomly selected uniformly from values other than the CF's observed value as the single estimate and $\{\bm{Q}^{(0)}_{(j)}\}_{j \in [F^c]}$ as the estimated confidence.
When IPAL was executed based on the proposed paper \cite{zhang2015solving}, the confidences of the CFs' exact values in the test data were not calculated, and only the CFs' estimated exact values were provided.
In this section, to compare the confidence propagation performance, IPAL's confidence propagation was conducted considering all data as training data.
The evaluation utilized both the single estimate and the estimated confidence obtained by IPAL.

First, using Tables \ref{tab:cf_score_bank} and \ref{tab:cf_score_adult}, we compare {\it comp} and IPAL to assess the effectiveness of the existing PLL method in estimating CFs' exact values. 
For all CFs, IPAL achieved higher Acc and F1 than {\it comp}. 
Thus, from the perspective of Acc and F1, IPAL is considered effective in estimating these exact values. 
However, compared to {\it comp}, IPAL exhibited worse CE and SE.
Therefore, when utilizing the estimated confidences of CFs' exact value for output label prediction, IPAL cannot be deemed an effective method.

In contrast, when comparing IPAL and {\it proposed}, {\it proposed} achieved higher Acc and F1 scores than IPAL for most CFs. 
Additionally, {\it proposed} significantly improved CE and SE compared to {\it comp}. 
As shown in Table \ref{tab:cf_score_bank}, for the Bank dataset, {\it proposed} demonstrated superior estimation performance than IPAL across all CFs and all evaluation metrics.
From Table \ref{tab:cf_score_adult}, it is evident that for the Adult dataset, {\it proposed} demonstrated outperformed than IPAL for six CFs, excluding education, and for most evaluation metrics. 
Particularly, {\it proposed} significantly enhanced SE for all CFs and notably improved CE for five CFs. 
Consequently, from the perspectives of Acc and F1, {\it proposed} has been experimentally shown to be effective in estimating CFs' exact values, outperforming IPAL. 
Furthermore, since {\it proposed} significantly improves CE and SE, it demonstrates a tendency to adequately reduce confidences for values other than the CFs' exact values.

However, Tables \ref{tab:cf_score_bank} and \ref{tab:cf_score_adult} reveal that {\it proposed} did not perform as well in accurately estimating the exact values for education, occupation, and native-country in the Adult dataset and job in the Bank dataset.
For occupation, native-country and job, {\it proposed} achieved only minor improvements, whereas for education, it exhibited a decrease in both Acc and F1 compared to {\it comp}.
Additionally, the reduction in SE by the proposed method was relatively small for these features.
These features are characterized by a high number of unique values, whereas those for which {\it proposed} performed best typically have fewer unique values.
Consequently, our proposed method tends to yield more accurate estimations for CFs with a smaller number of unique values.
Specifically, as the number of unique values for a CF increases, fulfilling Assumption \ref{asm:smoothness} becomes more challenging, potentially hindering the improvement of estimation accuracy for certain CFs.

This discussion provides guidance for the practical application of the proposed method.
In practice, it is difficult to evaluate how well the proposed method performs in estimating CFs' exact values, as these cannot be directly obtained.
Conversely, we observed that when the number of unique values for each CF is large, or when the decrease in SE is small, estimation accuracy tends to be low.
This suggests that estimation accuracy can be inferred to some extent based on the number of unique values of a CF and the reduction in SE.
Using the observed values as they are can be interpreted as employing the minimum necessary information without errors.
In other words, for certain CFs where exact values cannot be reliably estimated, it may be necessary to use the observed values directly.
Therefore, it may be effective to determine whether to use the estimated results of CFs' exact values or the observed values based on the number of unique values and the reduction in SE.

\subsection{Evaluation of Prediction Quality for the Target Feature}
\label{subsec:exp_predict_target}

\begin{table*}[t]
    \caption{The predictive accuracy of output labels using the estimated results of the proposed method (Bank)}
    \centering
    \begin{tabular}{l||l|l|l|l} 
    \hline
    Method & LR & RF & AdaBoost & MLP\\ 
    \hline \hline 
    ord & 0.4060 ±0.0075 & 0.4654 ±0.0094 & 0.4511 ±0.0041 & 0.4548 ±0.0311 \\ 
    \hline
    comp & 0.2586 ±0.0080 & 0.4047 ±0.0085 & 0.4175 ±0.0132 & 0.4198 ±0.0401 \\ 
    IPAL & 0.2709 ±0.0079& {\bf 0.4431 ±0.0078} & 0.4298 ±0.0132 & 0.4076 ±0.0255 \\ 
    proposed (soft)& {\bf 0.3279 ±0.0261} & 0.3488 ±0.0119 & 0.3539 ±0.0075& {\bf 0.4543 ±0.0280} \\ 
    proposed (hard) & 0.2939 ±0.0180 & 0.4335 ±0.0113& {\bf 0.4357 ±0.0114} & 0.4012 ±0.0622 \\ 
    \hline
    \end{tabular}
    \label{tab:pred_score_bank}
\end{table*}

\begin{table*}[t]
    \caption{The predictive accuracy of output labels using the estimated results of the proposed method (Adult)}
    \centering
    \begin{tabular}{l||l|l|l|l} 
    \hline
    Method & LR & RF & AdaBoost & MLP\\ 
    \hline \hline 
    ord & 0.6534 ±0.0058 & 0.6745 ±0.0085 & 0.6448 ±0.0093 & 0.6617 ±0.0122 \\ 
    \hline
    comp & 0.4243 ±0.0075 & {\bf 0.5834 ±0.0086} & {\bf 0.5647 ±0.0096} & {\bf 0.5706 ±0.0391} \\ 
    IPAL & 0.4220 ±0.0178 & 0.5139 ±0.0079 & 0.5306 ±0.0088 & 0.4401 ±0.0542 \\ 
    proposed (soft)& {\bf 0.4746 ±0.0135} & 0.4463 ±0.0118 & 0.4352 ±0.0073 & 0.4435 ±0.0528 \\ 
    proposed (hard) & 0.4025 ±0.0111 & 0.5184 ±0.0046 & 0.5259 ±0.0098 & 0.4425 ±0.0382 \\ 
    \hline
    \end{tabular}
    \label{tab:pred_score_adult}
\end{table*}

\begin{table*}[t]
    \caption{The predictive accuracy of output labels when only certain CFs are estimated using the proposed method (Adult)}
    \centering
    \begin{tabular}{l||l|l|l|l} 
    \hline
    Method & LR & RF & AdaBoost & MLP\\ 
    \hline \hline 
    ord & 0.6534 ±0.0058 & 0.6745 ±0.0085 & 0.6448 ±0.0093 & 0.6617 ±0.0122 \\ 
    \hline
    comp & 0.4243 ±0.0075 & 0.5834 ±0.0086 & 0.5647 ±0.0096 & 0.5706 ±0.0391 \\ 
    IPAL & 0.4360 ±0.0160 & 0.5966 ±0.0064 & 0.5718 ±0.0059 & 0.5762 ±0.0120 \\ 
    proposed (soft)& {\bf 0.4623 ±0.0169} & 0.5530 ±0.0051 & 0.5184 ±0.0073& {\bf 0.5805 ±0.0250} \\ 
    proposed (hard) & 0.4256 ±0.0094& {\bf 0.5988 ±0.0041}& {\bf 0.5792 ±0.0071} & 0.5780 ±0.0184 \\ 
    \hline
    \end{tabular}
    \label{tab:pred_score_adult_revise}
\end{table*}

In this section, we evaluated the predictive performance for output labels using CFs' estimated exact values by our proposed method as part of the input.
We employed F1 as the evaluation metric, presenting averages and standard deviations across five trials.
The hyperparameters for our proposed method were set as follows: $T=100, k=20, \gamma=0.25$.
For IPAL, the hyperparameters were set as: $T=100, k=20, \alpha'=0.90$.
We utilized logistic regression (LR), random forest (RF), AdaBoost, and MultiLayer Perceptron (MLP) as learning algorithms for training the label prediction model $p_{\bm{\theta}}$.
The MLP architecture consisted of four layers with widths of 100, 200, 200, and 100 neurons, respectively, starting from the input layer.
AdaBoost employed decision stumps as weak learners.
LR and MLP are optimized with $J_{\mathrm{KL}}$ as the objective function, whereas RF and Adaboost are optimized with other objective functions.
RF and Adaboost were selected for comparison as they can utilize the estimation results from our proposed method as inputs.

In this section, we compare five methods: {\it ord}, {\it comp}, {\it proposed (soft)}, {\it proposed (hard)}, and IPAL.
{\it ord} denotes ordinary learning where CFs' exact values are used as part of inputs.
{\it comp} represents the case where the CFs' observed values are used directly as part of inputs.
{\it proposed (soft)} refers to using the confidences of CFs' exact values estimated by our proposed method as part of inputs.
{\it proposed (hard)} refers to using the value with the highest estimated confidence as part of inputs.
For IPAL, following the approach described in \cite{zhang2015solving}, CFs' exact values in the training data were estimated using confidence propagation.
These estimates were subsequently used to estimate CFs' exact values in the test data without further confidence propagation. 
The training and test data containing the resulting estimates were then used to train $p_{\bm{\theta}}$ and predict output labels, respectively.

Table \ref{tab:pred_score_bank} presents the evaluation results on the Bank dataset. 
In the Bank dataset, IPAL outperforms {\it comp} across the three learning algorithms: LR, RF, and AdaBoost.
In contrast, {\it proposed (soft)} or {\it proposed (hard)} outperforms {\it comp} in all learning algorithms and achieves higher predictive accuracy that is equal to or greater than that of IPAL in the three learning algorithms: LR, AdaBoost, and MLP.
When using RF, {\it proposed} showed slightly inferior results compared to existing methods; however, under other conditions, it demonstrated strong predictive performance. 
In particular, {\it proposed (soft)} + MLP produced results comparable to the standard learning approach, {\it ord}.
Thus, for the Bank dataset, our proposed method was demonstrated to have a high potential for improving the prediction accuracy of output labels while accurately estimating the exact values of CFs.

Table \ref{tab:pred_score_adult} presents the evaluation results for the Adult dataset. 
In the Adult dataset, for learning algorithms other than LR, {\it proposed} underperformed {\it comp} and IPAL. 
Additionally, IPAL's performance was lower than that of {\it comp}. 
This outcome is likely attributed to the low quality of estimated exact values for multiple CFs in the Adult dataset, as indicated in Table \ref{tab:cf_score_adult}.
In other words, using estimated exact values of difficult-to-estimate CFs likely compromised the quality of the training data, making the learning process more challenging.

To investigate this further, we conducted an experiment in which we did not estimate the exact values of CFs with low estimation quality.
Instead, we used the observed values of these CFs for training. 
Table \ref{tab:pred_score_adult_revise} presents the results when education, occupation, and native-country in the Adult dataset were selected as the CFs not to be estimated by either {\it proposed} and IPAL. 
These CFs were chosen because their estimated exact values demonstrated lower quality, as shown in Table \ref{tab:cf_score_adult}. 
Comparing Tables \ref{tab:pred_score_adult} and \ref{tab:pred_score_adult_revise}, it is evident that the latter exhibits better performance across most learning algorithms when using either IPAL or {\it proposed}. 
Thus, the experiment confirmed that enforcing the estimation of exact values for certain difficult-to-estimate CFs resulted in decreased prediction accuracy when using either IPAL or {\it proposed}.
Furthermore, Table \ref{tab:pred_score_adult_revise} shows that using {\it proposed} achieved better predictive performance that is equal to or greater than that of both {\it comp} and IPAL for all learning algorithms. 
Also, identifying which CFs to apply the proposed method to is feasible in practical applications. 
Based on our earlier discussion, the CFs to which the proposed method should be applied can be determined by considering the number of unique values of each CF and the extent of SE reduction in confidence achieved by the proposed method.

Based on these results, the decision of whether to use {\it (soft)} or {\it (hard)} in {\it proposed} is determined as follows. 
From Tables \ref{tab:pred_score_bank} and \ref{tab:pred_score_adult_revise}, it is evident that {\it proposed (soft)} perform better for LR and MLP, whereas {\it proposed (hard)} is more effective for RF and AdaBoost.
{\it (soft)} directly inputs the estimated confidence into the label prediction model, allowing the model to account for the uncertainty of the estimation.
In contrast, {\it (hard)} selects a single value based on this estimated confidence, simplifying the input to the label prediction model by discarding uncertainty information.
Both LR and MLP are algorithms that construct prediction models through a combination of linear and nonlinear transformations. 
With these methods, it is considered that it was easy to optimize each transformation to utilize the confidence represented by a real-valued vector in {\it (soft)}.
Conversely, since AdaBoost uses decision stumps as weak learners, both RF and AdaBoost are decision tree-based methods. 
These algorithms primarily focus on optimizing the splits in decision trees; thus, {\it (hard)}, which employs a simplified single estimated value, is considered to facilitate this optimization compared to using the estimated confidence.
Therefore, it is advisable to use {\it proposed (hard)} for decision tree-based learning algorithms and {\it proposed (soft)} for other learning algorithms that effectively handle confidence.

Overall, the experimental results demonstrate that our proposed method is highly effective by selecting appropriate CFs for estimation and utilizing suitable learning algorithms for constructing label prediction models.
Specifically, for the CFs selected for estimation, our proposed method reduces the ambiguity regarding the exact values of CFs, enhancing predictive performance while ensuring interpretability.
Additionally, we anticipate that the selection of CFs for estimation will be straightforward in practical applications.

\subsection{Sensitivity Analysis}
\label{subsec:exp_sensitive_analysis}

\begin{figure}[t]
    \centering
    \includegraphics[width=\linewidth]{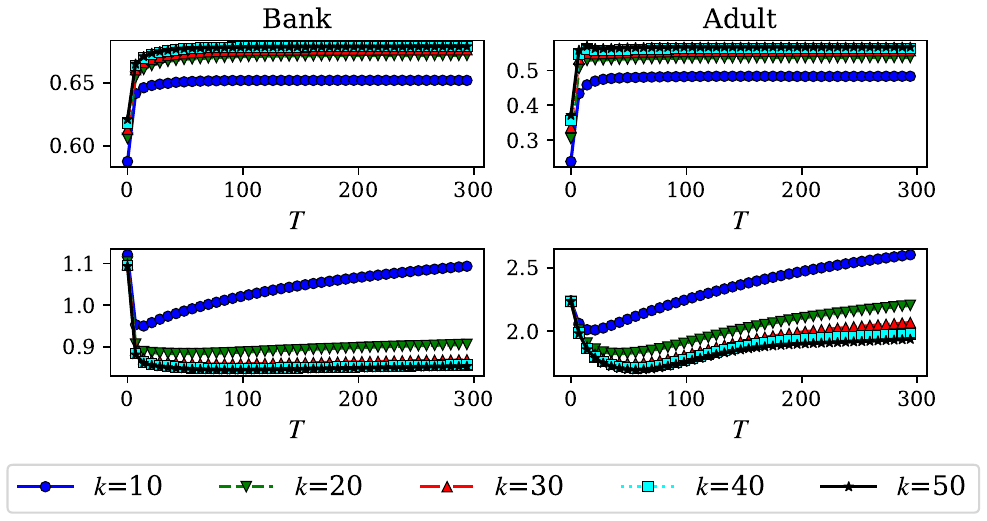}
    \caption{The relationship between $T$ and the estimation quality of CFs' exact values ($\gamma=0$, with Eq. \eqref{eq:correct_own_comp}).}
    \label{fig:convergence}
\end{figure}

\begin{table}[t]
    \caption{Relationship between Eq. \eqref{eq:correct_own_comp} and the estimation quality of CFs' exact values ($T=100$, $k=20$, $\gamma=0$).}
    \centering
    \begin{tabular}{l||c|c|c|c} 
    \hline
    Dataset & \multicolumn{2}{c|}{Bank} & \multicolumn{2}{c}{Adult} \\ 
\hline 
    & \multicolumn{1}{c|}{OFF} & ON& OFF & ON\\ 
    \hline \hline 
    Acc & \multicolumn{1}{c|}{0.61 ±0.00} & \multicolumn{1}{c|}{0.67 ±0.00} & \multicolumn{1}{c|}{0.57 ±0.00} & 0.53 ±0.00\\ 
    F1 & \multicolumn{1}{c|}{0.31 ±0.00} & \multicolumn{1}{c|}{0.50 ±0.00} & \multicolumn{1}{c|}{0.16 ±0.00} & 0.18 ±0.00\\ 
    CE & \multicolumn{1}{c|}{1.37 ±0.00} & \multicolumn{1}{c|}{0.89 ±0.00} & \multicolumn{1}{c|}{2.35 ±0.00} & 1.92 ±0.01\\ 
    SE & \multicolumn{1}{c|}{1.48 ±0.00} & \multicolumn{1}{c|}{0.73 ±0.00} & \multicolumn{1}{c|}{2.40 ±0.00} & 1.50 ±0.01\\ 
    \hline
    \end{tabular}
    \label{tab:correct_own_comp}
\end{table}

\begin{figure}[t]
    \centering
    \includegraphics[width=\linewidth]{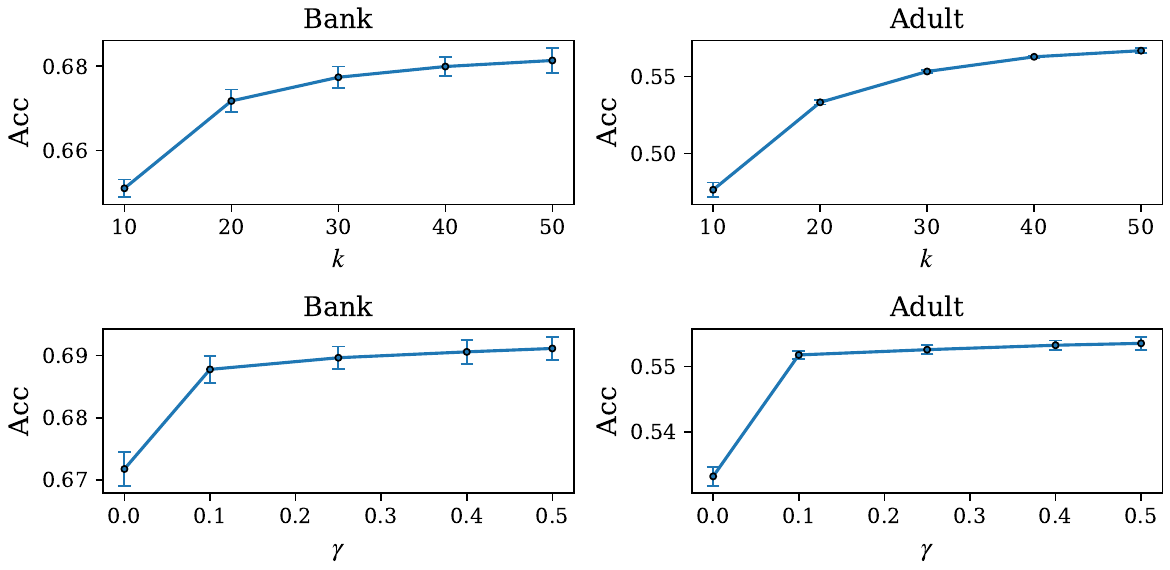}
    \caption{The relationship between each of $k$ and $\gamma$ and the estimation quality of CFs' exact values (with Eq. \eqref{eq:correct_own_comp}, $T=100$, $k=20$, $\gamma=0$ when their values are not varied).}
    \label{fig:vary_k_gamma}
\end{figure}

We investigated how the processing represented by Eq.\eqref{eq:correct_own_comp} and the hyperparameters of the proposed method $T$, $k$ and $\gamma$ affect the quality of estimating the exact values of CFs.
For the experiment on $T$, we present the results of a single trial, whereas all other experiments, show the aggregated results from five trials.
The experiments in this section evaluate the average scores of the estimated results for all CFs, with the proposed method applied using the entire dataset.

Figure \ref{fig:convergence} illustrates the number of iterations required for the proposed method to achieve convergence.
The figure indicates that convergence with respect to Acc is attained in fewer than $100$ iterations for each setting of $k$.
However, we observe an increase in the CE value in the middle of the process, particularly in the Adult dataset.
This phenomenon can be attributed to the challenges in satisfying Assumption \ref{asm:smoothness}, making it more likely for incorrect confidences propagating after a certain number of iterations.
Since the Adult dataset contains difficult-to-estimate CFs based on Section \ref{subsec:exp_predict_target}, a high number of iterations is likely to significantly worsen the estimating performance for their exact values.
Therefore, when dealing with difficult-to-estimate CFs, it is necessary to either reduce the number of iterations or avoid estimating those CFs.
Additionally, it can be observed that this phenomenon diminishes as the value of $k$ increases. 
Therefore, it is considered desirable to set $k$ sufficiently large.

Table \ref{tab:correct_own_comp} summarizes the relationship between the processing described in Eq. \eqref{eq:correct_own_comp} and the quality of estimation for CFs' exact values.
Setting $\gamma=0$ indicates that no second round of confidence propagation using the estimated results of CFs' exact values from the first round is performed.
Table \ref{tab:correct_own_comp} shows that the process in Eq. \eqref{eq:correct_own_comp} improved F1, CE, and SE for both datasets.
The experimental results confirm that applying Eq. \eqref{eq:correct_own_comp} contributes to the reduction of $J_{\mathrm{SL}}$ due to improvements in CE.

Figure \ref{fig:vary_k_gamma} shows the relationship between each of $k$ and $\gamma$ and Acc in estimating CFs' exact values.
The results indicate that Acc tends to increase as the values of $k$ and $\gamma$ become larger within the range of this experiment.
Regarding $k$, this is attributed to the consideration of a sufficient number of neighboring points, which improves weight optimization and, consequently, results in higher accuracy in these datasets.
The result of $\gamma$ confirms that applying the proposed method to a combination of OFs' values and CFs' estimated exact values is effective.

From these experiments, three key points emerged regarding the datasets.
First, the proposed method converges within $100$ iterations in the two datasets, making it practical for implementation.
Second, the processing described in Eq. \eqref{eq:correct_own_comp} is effective in terms of CE and SE.
Third, larger values of $k$ and $\gamma$ are particularly beneficial for the performance of our proposed method.
Additionally, we have confirmed that applying the proposed method to a combination of OFs' values and CFs' estimated exact values is effective.

\section{Conclusion}
\label{sec:conclusion}

Due to the impracticality of obtaining high-precision observations and the need for privacy protection, qualitative input features may contain complementary rather than precise information.
However, learning scenarios addressing this situation have not been previously discussed.
In this study, we proposed a novel learning scenario called CFL for learning from data that include CFs defined by complementary information indicating ``what it is not.''
First, we derived an objective function, from an information-theoretic perspective for predicting output labels based on inputs containing the estimated results of CFs' exact values.
We demonstrated that this objective function serves as an upper bound for the objective function used in ordinary learning.
Next, we proposed a graph-based iterative method for estimating CFs' exact values based on the derived objective function. 
To establish the theoretical validity of this method, we first developed a method that theoretically guarantees improved confidences in CFs' exact values at each iteration under the hypothetical setting that CFs' exact values can be self-referenced.
Subsequently, we formulated a practically feasible approximation of this method.
The results of numerical experiments using real-world data indicated that our proposed method yields highly accurate estimations for CFs with a small number of unique values.
Furthermore, we confirmed that appropriately combining our proposed method with various learning algorithms leads to exceptional predictive performance.

\section*{Acknowledgment}
This work was supported in part by the Japan Society for the Promotion of Science through Grants-in-Aid for Scientific Research (C) (23K11111).

\bibliographystyle{IEEEtran}
\bibliography{IEEEabrv,reference}

\begin{thebibliography}{10}
\providecommand{\url}[1]{#1}
\csname url@samestyle\endcsname
\providecommand{\newblock}{\relax}
\providecommand{\bibinfo}[2]{#2}
\providecommand{\BIBentrySTDinterwordspacing}{\spaceskip=0pt\relax}
\providecommand{\BIBentryALTinterwordstretchfactor}{4}
\providecommand{\BIBentryALTinterwordspacing}{\spaceskip=\fontdimen2\font plus
\BIBentryALTinterwordstretchfactor\fontdimen3\font minus \fontdimen4\font\relax}
\providecommand{\BIBforeignlanguage}[2]{{%
\expandafter\ifx\csname l@#1\endcsname\relax
\typeout{** WARNING: IEEEtran.bst: No hyphenation pattern has been}%
\typeout{** loaded for the language `#1'. Using the pattern for}%
\typeout{** the default language instead.}%
\else
\language=\csname l@#1\endcsname
\fi
#2}}
\providecommand{\BIBdecl}{\relax}
\BIBdecl

\bibitem{chapelle2006semi}
O.~Chapelle, B.~Scholkopf, and A.~Zien, ``Semi-supervised learning. 2006,'' \emph{Cambridge, Massachusettes: The MIT Press View Article}, vol.~2, 2006.

\bibitem{natarajan2013learning}
\BIBentryALTinterwordspacing
N.~Natarajan, I.~S. Dhillon, P.~K. Ravikumar, and A.~Tewari, ``Learning with noisy labels,'' in \emph{Advances in Neural Information Processing Systems}, C.~Burges, L.~Bottou, M.~Welling, Z.~Ghahramani, and K.~Weinberger, Eds., vol.~26.\hskip 1em plus 0.5em minus 0.4em\relax Curran Associates, Inc., 2013. [Online]. Available: \url{https://proceedings.neurips.cc/paper_files/paper/2013/file/3871bd64012152bfb53fdf04b401193f-Paper.pdf}
\BIBentrySTDinterwordspacing

\bibitem{cour2011learning}
T.~Cour, B.~Sapp, and B.~Taskar, ``Learning from partial labels,'' \emph{The Journal of Machine Learning Research}, vol.~12, pp. 1501--1536, 2011.

\bibitem{ishida2017cll}
T.~Ishida, G.~Niu, W.~Hu, and M.~Sugiyama, ``Learning from complementary labels,'' in \emph{Advances in Neural Information Processing Systems}, I.~Guyon, U.~V. Luxburg, S.~Bengio, H.~Wallach, R.~Fergus, S.~Vishwanathan, and R.~Garnett, Eds., vol.~30.\hskip 1em plus 0.5em minus 0.4em\relax Curran Associates, Inc., 2017.

\bibitem{ishida2019complementary}
T.~Ishida, G.~Niu, A.~Menon, and M.~Sugiyama, ``Complementary-label learning for arbitrary losses and models,'' in \emph{International Conference on Machine Learning}.\hskip 1em plus 0.5em minus 0.4em\relax PMLR, 2019, pp. 2971--2980.

\bibitem{morvan2020neumiss}
\BIBentryALTinterwordspacing
M.~Le~Morvan, J.~Josse, T.~Moreau, E.~Scornet, and G.~Varoquaux, ``Neumiss networks: differentiable programming for supervised learning with missing values.'' in \emph{Advances in Neural Information Processing Systems}, H.~Larochelle, M.~Ranzato, R.~Hadsell, M.~Balcan, and H.~Lin, Eds., vol.~33.\hskip 1em plus 0.5em minus 0.4em\relax Curran Associates, Inc., 2020, pp. 5980--5990. [Online]. Available: \url{https://proceedings.neurips.cc/paper_files/paper/2020/file/42ae1544956fbe6e09242e6cd752444c-Paper.pdf}
\BIBentrySTDinterwordspacing

\bibitem{morvan2021whats}
\BIBentryALTinterwordspacing
M.~Le~Morvan, J.~Josse, E.~Scornet, and G.~Varoquaux, ``What’s a good imputation to predict with missing values?'' in \emph{Advances in Neural Information Processing Systems}, M.~Ranzato, A.~Beygelzimer, Y.~Dauphin, P.~Liang, and J.~W. Vaughan, Eds., vol.~34.\hskip 1em plus 0.5em minus 0.4em\relax Curran Associates, Inc., 2021, pp. 11\,530--11\,540. [Online]. Available: \url{https://proceedings.neurips.cc/paper_files/paper/2021/file/5fe8fdc79ce292c39c5f209d734b7206-Paper.pdf}
\BIBentrySTDinterwordspacing

\bibitem{zhu2002learning}
X.~Zhu and Z.~Ghahramani, ``Learning from labeled and unlabeled data with label propagation,'' \emph{ProQuest number: information to all users}, 2002.

\bibitem{zhang2015solving}
M.-L. Zhang and F.~Yu, ``Solving the partial label learning problem: An instance-based approach.'' in \emph{IJCAI}, 2015, pp. 4048--4054.

\bibitem{lin2023reduction}
W.-I. Lin and H.-T. Lin, ``Reduction from complementary-label learning to probability estimates,'' in \emph{Pacific-Asia Conference on Knowledge Discovery and Data Mining}.\hskip 1em plus 0.5em minus 0.4em\relax Springer, 2023, pp. 469--481.

\bibitem{katsura20bridging}
\BIBentryALTinterwordspacing
Y.~Katsura and M.~Uchida, ``Bridging ordinary-label learning and complementary-label learning,'' in \emph{Proceedings of The 12th Asian Conference on Machine Learning}, ser. Proceedings of Machine Learning Research, S.~J. Pan and M.~Sugiyama, Eds., vol. 129.\hskip 1em plus 0.5em minus 0.4em\relax PMLR, 18--20 Nov 2020, pp. 161--176. [Online]. Available: \url{https://proceedings.mlr.press/v129/katsura20a.html}
\BIBentrySTDinterwordspacing

\bibitem{jin2002learning}
\BIBentryALTinterwordspacing
R.~Jin and Z.~Ghahramani, ``Learning with multiple labels,'' in \emph{Advances in Neural Information Processing Systems}, S.~Becker, S.~Thrun, and K.~Obermayer, Eds., vol.~15.\hskip 1em plus 0.5em minus 0.4em\relax MIT Press, 2002. [Online]. Available: \url{https://proceedings.neurips.cc/paper_files/paper/2002/file/653ac11ca60b3e021a8c609c7198acfc-Paper.pdf}
\BIBentrySTDinterwordspacing

\bibitem{nguyen-caruana2008classfication}
\BIBentryALTinterwordspacing
N.~Nguyen and R.~Caruana, ``Classification with partial labels,'' in \emph{Proceedings of the 14th ACM SIGKDD International Conference on Knowledge Discovery and Data Mining}, ser. KDD '08.\hskip 1em plus 0.5em minus 0.4em\relax New York, NY, USA: Association for Computing Machinery, 2008, p. 551^^e2^^80^^93559. [Online]. Available: \url{https://doi.org/10.1145/1401890.1401958}
\BIBentrySTDinterwordspacing

\bibitem{gong-liu2018regularization}
C.~Gong, T.~Liu, Y.~Tang, J.~Yang, J.~Yang, and D.~Tao, ``A regularization approach for instance-based superset label learning,'' \emph{IEEE Transactions on Cybernetics}, vol.~48, no.~3, pp. 967--978, 2018.

\bibitem{sun2020pp}
K.~Sun, Z.~Min, and J.~Wang, ``Pp-pll: Probability propagation for partial label learning,'' in \emph{Machine Learning and Knowledge Discovery in Databases: European Conference, ECML PKDD 2019, W{\"u}rzburg, Germany, September 16--20, 2019, Proceedings, Part II}.\hskip 1em plus 0.5em minus 0.4em\relax Springer, 2020, pp. 123--137.

\bibitem{zhou2003label-spreading}
\BIBentryALTinterwordspacing
D.~Zhou, O.~Bousquet, T.~Lal, J.~Weston, and B.~Sch\"{o}lkopf, ``Learning with local and global consistency,'' in \emph{Advances in Neural Information Processing Systems}, S.~Thrun, L.~Saul, and B.~Sch\"{o}lkopf, Eds., vol.~16.\hskip 1em plus 0.5em minus 0.4em\relax MIT Press, 2003. [Online]. Available: \url{https://proceedings.neurips.cc/paper_files/paper/2003/file/87682805257e619d49b8e0dfdc14affa-Paper.pdf}
\BIBentrySTDinterwordspacing

\bibitem{Dua:2019}
\BIBentryALTinterwordspacing
D.~Dua and C.~Graff, ``{UCI} machine learning repository,'' 2017. [Online]. Available: \url{http://archive.ics.uci.edu/ml}
\BIBentrySTDinterwordspacing

\end{thebibliography}

\appendix

\subsection{Proof of Theorem \ref{thm:objective_relation}}
\label{subsec:proof_obj_relation}

Because $Y \indep \widebar{\bm{X}}^c | \bm{X}$ holds, as shown in Figure \ref{fig:graphical_model}, the following transformation is valid:
\begin{align}
    & D_{\mathrm{KL}}(p_*(Y|\bm{X}) || p_{\bm{\theta}}(Y|\bm{X})) \notag \\
    &~~ = D_{\mathrm{KL}}(p_*(Y|\bm{X}, \widebar{\bm{X}}^c) || p_{\bm{\theta}}(Y|\bm{X}, \widebar{\bm{X}}^c)) \notag \\
    &~~ = - \mathbb{E}_{p_*(\bm{x}, \bar{\bm{x}}^c, y)}[ \log p_{\bm{\theta}}(y|\bm{x, \bar{\bm{x}}^c}) ] - \mathbb{H}(Y|\bm{X}).
    \label{eq:ord_KL_extension}
\end{align}
Here, $\mathbb{H}$ denotes entropy.
The first term on the RHS of Eq. \eqref{eq:ord_KL_extension} multiplied by $-1$ is transformed as follows:

\begin{align}
    &\mathbb{E}_{p_*(\bm{x}, \bar{\bm{x}}^c, y)}[ \log p_{\bm{\theta}}(y|\bm{x, \bar{\bm{x}}^c}) ] \notag \\
    &= \mathbb{E}_{p_*(\bm{x}, \bar{\bm{x}}^c, y) q_{\bm{\eta}}(\hat{\bm{x}}^c|\bar{\bm{x}})} \bigg[ \log \frac{p_{\bm{\theta}}(y|\bm{x}, \bar{\bm{x}}^c) p_*(\bm{x}^c|\bar{\bm{x}}) q_{\bm{\eta}}(\hat{\bm{x}}^c|\bar{\bm{x}})}{p_*(\bm{x}^c|\bar{\bm{x}}) q_{\bm{\eta}}(\hat{\bm{x}}^c|\bar{\bm{x}})} \bigg] \notag \\
    &\ge \mathbb{E}_{p_*(\bm{x}^{o}, \bar{\bm{x}}^c, y) q_{\bm{\eta}}(\hat{\bm{x}}^c|\bar{\bm{x}})} \bigg[ \log \frac{p_{\bm{\theta}}(y|\bar{\bm{x}}) q_{\bm{\eta}}(\hat{\bm{x}}^c|\bar{\bm{x}})}{q_{\bm{\eta}}(\hat{\bm{x}}^c|\bar{\bm{x}})} \bigg].
    \label{eq:obj_rel_proof_tmp1}
\end{align}

In the last inequality of Eq. \eqref{eq:obj_rel_proof_tmp1}, we use the log sum inequality for $\bm{x}^{c}$.
Here, $p_{*,\bm{\eta}}(\bm{x}^{o}, \bar{\bm{x}}^c, y, \hat{\bm{x}}^c) \equiv  p_*(\bm{x}^{o}, \bar{\bm{x}}^c, y) q_{\bm{\eta}}(\hat{\bm{x}}^c|\bar{\bm{x}}^c, \bm{x}^{o})$.
From Figure \ref{fig:graphical_model}, because $Y \indep \widehat{\bm{X}}^c |\widebar{\bm{X}}^c, \bm{X}^{o}$ holds, we define $r_{\bm{\eta},\bm{\theta}}(y,\hat{\bm{x}}^c | \bar{\bm{x}}^c, \bm{x}^{o})\equiv p_{\bm{\theta}}(y|\bar{\bm{x}}^c, \bm{x}^{o}) q_{\bm{\eta}}(\hat{\bm{x}}^c|\bar{\bm{x}}^c, \bm{x}^{o})$.
Then Eq. \eqref{eq:obj_rel_proof_tmp1} can be transformed as follows:
\begin{align}
    & \mathbb{E}_{p_*(\bm{x}, \bar{\bm{x}}^c, y)} [\log p_{\bm{\theta}}(y|\bm{x, \bar{\bm{x}}^c}) ] \notag \\
    &\ge \mathbb{E}_{p_{*,\bm{\eta}}(\bm{x}^{o}, \bar{\bm{x}}^c, y, \hat{\bm{x}}^c)} \bigg[ \log \frac{r_{\bm{\eta},\bm{\theta}}(y,\hat{\bm{x}}^c | \bar{\bm{x}}^c, \bm{x}^{o})}{q_{\bm{\eta}}(\hat{\bm{x}}^c|\bar{\bm{x}}^c, \bm{x}^{o})}  \bigg] \notag \\
    &= \mathbb{E}_{p_{*,\bm{\eta}}(\bm{x}^{o}, \bar{\bm{x}}^c, y, \hat{\bm{x}}^c)} \bigg[ \log \frac{ r_{\bm{\eta},\bm{\theta}}(y,\hat{\bm{x}}^c | \bar{\bm{x}}^c, \bm{x}^{o}) p(\bar{\bm{x}}^c|\bm{x}^{o}) }{ q_{\bm{\eta}}(\hat{\bm{x}}^c|\bar{\bm{x}}^c, \bm{x}^{o}) p(\bar{\bm{x}}^c|\bm{x}^{o}) }  \bigg]  \notag \\
    &\ge \mathbb{E}_{p_{*,\bm{\eta}}(\bm{x}^{o}, y, \hat{\bm{x}}^c)} \bigg[ \log \frac{r_{\bm{\eta},\bm{\theta}}(y,\hat{\bm{x}}^c |\bm{x}^{o})}{q_{\bm{\eta}}(\hat{\bm{x}}^c|\bm{x}^{o})} \bigg]  \notag \\
    &= \mathbb{E}_{p_{*,\bm{\eta}}(\bm{x}^{o}, y, \hat{\bm{x}}^c)} [\log p_{\bm{\theta}}(y|\hat{\bm{x}}^c, \bm{x}^{o}) ] \notag \\
    &= - \mathbb{E}_{p_{*,\bm{\eta}}(\bm{x}^{o}, y, \hat{\bm{x}}^c)} \bigg[ \log \frac{p_*(y|\hat{\bm{x}}^c, \bm{x}^{o})}{p_{\bm{\theta}}(y|\hat{\bm{x}}^c, \bm{x}^{o})} \bigg] \notag \\
    &~~~~~~~~ + \mathbb{E}_{p_{*,\bm{\eta}}(\bm{x}^{o}, y, \hat{\bm{x}}^c)} [ \log p_*(y|\hat{\bm{x}}^c, \bm{x}^{o}) ] \notag \\
    &= - D_{\mathrm{KL}}(p_*(Y|\widehat{\bm{X}}^c, \bm{X}^{o}) || p_{\bm{\theta}}(Y|\widehat{\bm{X}}^c, \bm{X}^{o})) \notag \\
    &~~~~~~~~~~~~~~~~~~~~~~ - \mathbb{H}(Y|\widehat{\bm{X}}^c, \bm{X}^{o}).
    \label{eq:obj_rel_proof_tmp2}
\end{align}

Here, $p(\bar{\bm{x}}^c|\bm{x}^{o}) \equiv \sum_{\bm{x}^{c} \in \mathcal{X}^c} \bar{p}(\bar{\bm{x}}^c|\bm{x}^c) p_*(\bm{x}^c|\bm{x}^{o})$, 
$p_{*,\bm{\eta}}(\bm{x}^{o}, y, \hat{\bm{x}}^c) \equiv \sum_{\bar{\bm{x}}^{c} \in \widebar{\mathcal{X}}^c} p_{*,\bm{\eta}}(\bm{x}^{o}, \bar{\bm{x}}^c, y, \hat{\bm{x}}^c)$, 
$r_{\bm{\eta},\bm{\theta}}(y,\hat{\bm{x}}^c |\bm{x}^{o}) \equiv \sum_{\bar{\bm{x}}^{c} \in \widebar{\mathcal{X}}^c} r_{\bm{\eta},\bm{\theta}}(y,\hat{\bm{x}}^c | \bar{\bm{x}}^c, \bm{x}^{o}) p(\bar{\bm{x}}^c|\bm{x}^{o})$ 
and $q_{\bm{\eta}}(\hat{\bm{x}}^c|\bm{x}^{o}) \equiv \sum_{\bar{\bm{x}}^{c} \in \widebar{\mathcal{X}}^c} q_{\bm{\eta}}(\hat{\bm{x}}^c|\bar{\bm{x}}^c, \bm{x}^{o}) p(\bar{\bm{x}}^c|\bm{x}^{o})$.

In the second inequality of Eq. \eqref{eq:obj_rel_proof_tmp2}, we apply the log sum inequality for $\bar{\bm{x}}^{c}$.

Therefore, substituting Eq. \eqref{eq:obj_rel_proof_tmp2} into Eq. \eqref{eq:ord_KL_extension} yields the following inequality, which proves Theorem \ref{thm:objective_relation}:

\begin{align}
    &D_{\mathrm{KL}}(p_*(Y|\bm{X}) || p_{\bm{\theta}}(Y|\bm{X})) \notag \\
    &~~~\le D_{\mathrm{KL}}(p_*(Y|\widehat{\bm{X}}^c, \bm{X}^{o}) || p_{\bm{\theta}}(Y|\widehat{\bm{X}}^c, \bm{X}^{o})) \notag \\
    &~~~~~~~~~~~~~~~ + \mathbb{I}(Y|\bm{X}^c, \bm{X}^{o}) - \mathbb{I}(Y|\widehat{\bm{X}}^c, \bm{X}^{o}). \notag
\end{align}

\subsection{Proof of Theorem \ref{thm:J_MI}}
\label{subsec:proof_J_MI}

Regarding $\mathbb{I}(Y,\widehat{\bm{X}}^{c}|\bm{X}^{o})$ defined by Eq. \eqref{eq:MI_hat}, from the definition of $p_{*,\bm{\eta}}(y,\hat{\bm{x}}^{c}|\bm{x}^{o})$ and $q_{\bm{\eta}}(\hat{\bm{x}}^{c}|\bm{x}^{o})$, the following holds:

\begin{align}
    &\mathbb{I}(Y,\widehat{\bm{X}}^{c}|\bm{X}^{o}) 
    = \mathbb{E}_{q_{\bm{\eta}}(\hat{\bm{x}}^{c}|\bm{x}^{o})p_*(\bm{x}^{o},y)}\bigg[\log \frac{p_{*,\bm{\eta}}(y, \hat{\bm{x}}^{c}|\bm{x}^{o})}{p_*(y|\bm{x}^{o})q_{\bm{\eta}}(\hat{\bm{x}}^{c}|\bm{x}^{o})}  \bigg] \notag \\
    &= \mathbb{E}_{q_{\bm{\eta}}(\hat{\bm{x}}^{c}|\bm{x}^{o})p_*(\bm{x}^{o},y)}\bigg[ \notag \\
    &~~~~~ \log \frac{  \mathbb{E}_{\bar{p}(\bar{\bm{x}}^{c}|\bm{x}^{c})p_*(\bm{x}^{c}|\bm{x}^{o})}[p_*(y|\bm{x}^{c},\bm{x}^{o}, \bar{\bm{x}}^{c})q_{\bm{\eta}}(\hat{\bm{x}}^{c}|\bm{x}^{o}, \bar{\bm{x}}^{c})]  }{p_*(y|\bm{x}^{o}) \mathbb{E}_{\bar{p}(\bar{\bm{x}}^{c}|\bm{x}^c)p_*(\bm{x}^c|\bm{x}^{o})}[q_{\bm{\eta}}(\hat{\bm{x}}^{c} | \bar{\bm{x}}^{c}, \bm{x}^{o})]}  \bigg] \notag \\
    &\le \mathbb{E}_{\bar{p}(\bar{\bm{x}}^{c}|\bm{x}^{c})p_*(\bm{x}^{c}|\bm{x}^{o})q_{\bm{\eta}}(\hat{\bm{x}}^{c}|\bm{x}^{o})p_*(\bm{x}^{o},y)}\bigg[ \notag \\
    &~~~~~~~~~~~~~~~ \log \frac{p_*(y|\bm{x}^{c}, \bm{x}^{o}, \bar{\bm{x}}^{c})q_{\bm{\eta}}(\hat{\bm{x}}^{c} | \bm{x}^{o}, \bar{\bm{x}}^{c}) }{p_*(y|\bm{x}^{o})q_{\bm{\eta}}(\hat{\bm{x}}^{c} | \bm{x}^{o}, \bar{\bm{x}}^{c})} \bigg] \notag \\
    &= \mathbb{E}_{p_*(\bm{x}^{c}|\bm{x}^{o})p_*(\bm{x}^{o},y)}\bigg[\log \frac{p_*(y|\bm{x}^{c}, \bm{x}^{o})}{p_*(y|\bm{x}^{o})} \bigg] \notag \\
    &= \mathbb{E}_{p_*(\bm{x}^{c}, \bm{x}^{o}, y)}\bigg[\frac{p_*(y, \bm{x}^{c}|\bm{x}^{o})}{p_*(y|\bm{x}^{o})p_*(\bm{x}^{c}|\bm{x}^{o})}\bigg] =\mathbb{I}(Y,\bm{X}^{c}|\bm{X}^{o}). \notag
\end{align}

In the inequality on the third line, the log-sum inequality was used. 
Consequently, Eq. \eqref{eq:J_MI_lower_bound} has been demonstrated.

From Eqs \eqref{eq:J_MI_lower_bound}, \eqref{eq:MI_star}, and \eqref{eq:MI_hat}, $J_{\mathrm{MI}}$ is minimized when $p_*(\bm{x}^{c}|\bm{x}^{o}) = q_{\bm{\eta}}(\bm{x}^{c}|\bm{x}^{o}), \forall \bm{x}^{c} \in \mathcal{X}^{c}, \forall \bm{x}^{o} \in \mathcal{X}^{o}$ holds. 
Therefore, the following holds:
\begin{align}
        D_{\mathrm{KL}}(p_*(\bm{X}^{c}|\bm{X}^{o}) || q_{\bm{\eta}}(\bm{X}^{c}|\bm{X}^{o})) =0 \Rightarrow J_{\mathrm{MI}} = 0. \notag
\end{align}

\subsection{Proof of Theorem \ref{thm:iterative_revision}}
\label{subsec:proof_iterative_revision}

For each $t\in [T]$, we define a manifold $\mathcal{M}_{t} = \{\sum^n_{i'=1}h_{i'}q^{(t-1)}(\bm{x}^{c}|\bm{x}_{i'}^{o}) | \sum^n_{i'=1}h_{i'}=1 \land (h_{i'}\ge 0, \forall i' \in [n]) \}$ on the probability distribution space of $\bm{X}^{c}$ by the linear mixture of $\{q^{(t-1)}(\bm{x}^{c}|\bm{x}_{i'}^{o})\}_{i'\in [n]}$.
Because the joint probability distribution on $\bm{X}^c$ is a discrete distribution, it belongs to an exponential family of probability distributions.
Thus, for any $t \in [T]$, $\mathcal{M}_{t}$ is $m$-flat.
For any $i\in [n]$, optimizing $\widehat{\bm{H}}_i$ using Eq. \eqref{eq:weight_opt_ideal} to obtain $q^{(t)}(\bm{x}^{c}|\bm{x}^{o}_i)$ is equivalent to an $e$-projection from $p_*(\bm{x}^{c}|\bm{x}^{o}_i)$ onto $\mathcal{M}_{t}$.
Therefore, since $q^{(t-1)}(\bm{x}^{c}|\bm{x}^{o}_i)$ is a point on $\mathcal{M}_{t}$, the following inequality holds for any $i \in [n]$ according to the Generalized Pythagorean Theorem:
\begin{align}
    &D_{\mathrm{KL}}(p_*(\bm{X}^{c}|\bm{x}^{o}_i) ||  q^{(t-1)}(\bm{X}^{c}|\bm{x}^{o}_i)) \notag \\
    &~~~~~= D_{\mathrm{KL}}(p_*(\bm{X}^{c}|\bm{x}^{o}) ||  q^{(t)}(\bm{X}^{c}|\bm{x}^{o})) \notag \\
    &~~~~~~~~ + D_{\mathrm{KL}}(q^{(t)}(\bm{X}^{c}|\bm{x}^{o}) ||  q^{(t-1)}(\bm{X}^{c}|\bm{x}^{o})) \notag \\
    &~~~~~\ge D_{\mathrm{KL}}(p_*(\bm{X}^{c}|\bm{x}^{o}) ||  q^{(t)}(\bm{X}^{c}|\bm{x}^{o})). \notag
\end{align}   
Hence, Theorem \ref{thm:iterative_revision} has been proven.

\subsection{Proof of Theorem \ref{thm:joint_marginal_equality}}
\label{subsec:proof_joint_marginal_equality}

First, we demonstrate that Eq. \eqref{eq:joint_marginal_equality} holds for any $i\in [n], j \in [F^{c}]$ and any $x^{c}_j \in \mathcal{X}^{c}_j$ when $t=0$.
For any $i\in [n], j \in [F^{c}]$, when $x^{c}_j \neq \bar{x}^{c}_{ij}$, regarding $\bm{x}^{c}_{\setminus j} \in \mathcal{X}^{c}_{\setminus j}$, the number of $\bm{x}^{c}_{\setminus j}$ that satisfies $q^{(0)}(\bm{x}^{c}_{\setminus j}, x^{c}_j | \bm{x}^{o}_i) \neq 0$ is $\prod^{F^c}_{j'=1, j'\neq j}(|\mathcal{X}^{c}_{j'}|-1)$.
Therefore, the following holds:

\begin{align}
    \sum_{\bm{x}^c_{\setminus j} \in \mathcal{X}^{c}_{\setminus j}}q^{(0)}(\bm{x}^c_{\setminus j}, x^{c}_j|\bm{x}^{o}_i)
    &= \frac{\prod^{F^c}_{\substack{j'=1,\\ j'\neq j}}(|\mathcal{X}^{c}_{j'}|-1)}{\prod^{F^c}_{j=1}(|\mathcal{X}^{c}_{j}|-1)} 
    = \frac{1}{|\mathcal{X}^{c}_{j'}|-1}. \notag
\end{align}
On the other hand, for any $i\in [n], j \in [F^{c}]$, when $x^{c}_j = \bar{x}^{c}_{ij}$, because $q^{(0)}(\bm{x}^{c}_{\setminus j}, x^{c}_j|\bm{x}^{c}_i)=0$ holds for any $\bm{x}^{c}_{\setminus j} \in \mathcal{X}^{c}_{\setminus j}$, the following holds:
\begin{align}
    \sum_{\bm{x}^c_{\setminus j} \in \mathcal{X}^{c}_{\setminus j}}q^{(0)}(\bm{x}^c_{\setminus j}, x^{c}_j|\bm{x}^{o}_i)
    = \begin{cases}
        \frac{1}{|\mathcal{X}^c_j|-1} & \text{if } x^{c}_j \neq \bar{x}^c_{ij} \\
        0 & \text{if } x^{c}_j = \bar{x}^c_{ij}
      \end{cases}. \notag
\end{align}
Hence, when $t=0$, for any $i\in [n], j \in [F^{c}]$ and any $x^{c}_j \in \mathcal{X}^{c}_j$, Eq. \eqref{eq:joint_marginal_equality} holds.

Next, assuming that Eq. \eqref{eq:joint_marginal_equality} holds for any $i\in [n], j \in [F^{c}]$ and any $x^{c}_j \in \mathcal{X}^{c}_j$ when $t=k \in \{1,\dots,T-1\}$, and the following holds for the case when $t=k+1$:
\begin{align}
\begin{split}
    &\sum_{\bm{x}^c_{\setminus j} \in \mathcal{X}^{c}_{\setminus j}}q^{(k+1)}(\bm{x}^c_{\setminus j}, x^{c}_j|\bm{x}^{o}_i) \notag \\
    &= \sum^n_{i'=1,i'\neq i}\hat{H}_{ii'} \sum_{\bm{x}^c_{\setminus j} \in \mathcal{X}^{c}_{\setminus j}} q^{(k)}(\bm{x}^c_{\setminus j}, x^{c}_j|\bm{x}^{o}_{i'}) \notag \\
    &=  \sum^n_{i'=1,i'\neq i}\hat{H}_{ii'}q^{(k)}(x^{c}_j|\bm{x}^{o}_{i'})  
    = q^{(k+1)}(x^{c}_j|\bm{x}^{o}_i).
\end{split}
\end{align}
The first and last equations use Eq. \eqref{eq:conf_prop_joint_detail} and Eq. \eqref{eq:conf_prop_marginal_detail}, which describe the propagation of joint confidence and marginal confidence element-wise, respectively. 
Therefore, when $t=k+1$, Eq. \eqref{eq:joint_marginal_equality} holds for any $i\in [n], j \in [F^{c}]$ and any $x^{c}_j \in \mathcal{X}^{c}_j$.
From the above, Theorem \ref{thm:joint_marginal_equality} is proven by mathematical induction.

\end{document}